\documentclass[journal, compsoc]{IEEEtran}
\usepackage[numbers, sort&compress]{natbib}
\usepackage{mathrsfs}
\usepackage{amsfonts}
\usepackage{times}
\usepackage{diagbox}
\usepackage{soul}
\usepackage{url}
\usepackage[utf8]{inputenc}
\usepackage{graphicx}
\usepackage{amsmath}
\usepackage{amsthm}
\usepackage{tabularx,booktabs}
\newcolumntype{Y}{>{\centering\arraybackslash}X}
\usepackage{multirow}
\usepackage{mathtools}
\usepackage{xcolor}
\usepackage{algorithmic}
\usepackage{subfigure}
\usepackage{enumitem}
\usepackage{tabularx}
\usepackage{relsize}
\usepackage{array}
\usepackage{multirow}
\usepackage{tikz}
\usetikzlibrary{decorations.pathreplacing,calligraphy}

\newcolumntype{?}{!{\vrule width 0.8pt}}
\setlist{nolistsep}

\newcommand\Tstrut{\rule{0pt}{2.3ex}}         
\newcommand\Bstrut{\rule[-1.3ex]{0pt}{0pt}}   

\begin{document}

\title{Enabling Energy-Efficient Deployment of\\ Large Language Models on Memristor Crossbar: A Synergy of Large and Small}  

\author{Zhehui~Wang, Tao~Luo*, Cheng~Liu, Weichen~Liu, Rick~Siow~Mong~Goh, Weng-Fai~Wong
 \thanks{Published in IEEE TPAMI}
 \thanks{*Corresponding author: Tao Luo, E-mail:
tluo001@e.ntu.edu.sg}
 \thanks{Zhehui Wang (E-mail: zhehui@connect.ust.hk), Tao Luo and Rick Siow Mong Goh are with the Institute of High Performance Computing (IHPC), Agency for Science,  Technology and Research (A*STAR), \#16-16 Connexis, 1 Fusionopolis Way, Singapore 138632, Republic of Singapore}
\thanks{Cheng Liu is with the Chinese Academy of Sciences.}
\thanks{Weichen Liu is with the Nanyang Technological University, Singapore.}
\thanks{Weng-Fai Wong is with the National University of Singapore.}
\thanks{DOI Bookmark: 10.1109/TPAMI.2024.3483654}
\vspace{-15pt}}

\IEEEtitleabstractindextext{
\begin{abstract}
Large language models (LLMs) have garnered substantial attention due to their promising applications in diverse domains. Nevertheless, the increasing size of LLMs comes with a significant surge in the computational requirements for training and deployment. Memristor crossbars have emerged as a promising solution, which demonstrated a small footprint and remarkably high energy efficiency in computer vision (CV) models. Memristors possess higher density compared to conventional memory technologies, making them highly suitable for effectively managing the extreme model size associated with LLMs.
However, deploying LLMs on memristor crossbars faces three major challenges. Firstly, the size of LLMs increases rapidly, already surpassing the capabilities of state-of-the-art memristor chips. Secondly, LLMs often incorporate multi-head attention blocks, which involve non-weight stationary multiplications that traditional memristor crossbars cannot support. Third, while memristor crossbars excel at performing linear operations, they are not capable of executing complex nonlinear operations in LLM such as softmax and layer normalization.
To address these challenges, we present a novel architecture for the memristor crossbar that enables the deployment of state-of-the-art LLM on a single chip or package, eliminating the energy and time inefficiencies associated with off-chip communication. 
Our testing on BERT$_{\textrm{Large}}$ showed negligible accuracy loss. Compared to traditional memristor crossbars, our architecture achieves enhancements of up to $39\times$ in area overhead and $18\times$ in energy consumption. Compared to modern TPU/GPU systems, our architecture demonstrates at least a $68\times$ reduction in the area-delay product and a significant 69\% energy consumption reduction.
\end{abstract}

}

\maketitle

\IEEEraisesectionheading{\section{Introduction}}

\IEEEPARstart{L}{arge} language models (LLMs), such as ChatGPT, LLaMA and PaLM, have become increasingly popular in recent years due to their ability to leverage vast amounts of professional knowledge by fine-tuning the model. This potential has been demonstrated in various domains, including medical and finance technologies. However, the size of LLMs has been growing rapidly with their increasing accuracy, resulting in huge computational complexity. Even inferring LLMs require powerful computing systems that consume a significant amount of energy. For example, the inference of ChatGPT requires approximately eight A100 GPU cards~\cite{Nvidia}, which poses challenges for local deployment, particularly in mobile environments. The OpenAI data center for ChatGPT consumes 23 million kWh based on monthly requests~\cite{khowaja2023chatgpt}. 
This high energy consumption and cost for model inference could constrain its further development, particularly as models become even larger. Thus, there is a significant demand for small-size LLM accelerators that can perform model inference more efficiently and cost-effectively.

Memristor crossbars are widely considered strong competitors for traditional machine learning accelerators~\cite{sebastian2020memory}. Numerous memristor crossbar systems have been proposed~\cite{ielmini2018memory,nambiar20200,choong2022hardware,luo2021nc,yang2022coreset}, showcasing low power consumption and low latency compared to classical accelerators. By taking advantage of the physical characteristics of memristive storage technology, the \textit{analog computation} can be performed using memristors~\cite{shafiee2016isaac,luo2016racetrack,luo2017novel}, which greatly boosts the accelerator's performance.
Different types of memristors using various \textit{Non-volatile memory} (NVM) technologies~\cite{chen2016review} exist, including resistive random access memory (RRAM), phase-change memory (PCM), spin-transfer torque magnetic random access memory (STT-RAM), and Flash memory. All of them are promising candidates for building high-efficiency accelerators. For the purposes of this discussion, we will use RRAM memristors as a representative of these NVM technologies.

Compared with the current leading memory technologies, such as DRAM (\textit{dynamic random-access memory}) and SRAM (\textit{static random-access memory}), memristors have higher density. For instance, each RRAM memristor occupies only 4$f^2$~\cite{zahoor2020resistive} area, where $f$ refers to the feature size of the chip. Considering the use of 4-bit memristors, a footprint of 274 $mm^2$ memory cells area is sufficient to store all 175 billion parameters of GPT-3, assuming the 14 nm technology. In contrast, the DRAM and SRAM require 6$f^2$~\cite{zahoor2020resistive} and 100$f^2$~\cite{zahoor2020resistive} for each bit of information. As a comparison, they demand significantly larger areas of 1646 $mm^2$ and 27440 $mm^2$ as memory cell area for GPT-3, assuming the same 14 nm technology node. By achieving a substantial decrease in physical size, there is a high chance to store the whole neural network models within a single chip or package, thus effectively eliminating the inefficiencies of off-chip communication in terms of both time and energy~\cite{oh20095}.
 
Despite the benefits of memristor-based machine learning accelerators and their wide applications in neural networks, it is still difficult to directly deploy LLMs on memristor-based accelerators due to three major challenges that constrain their usage.
\begin{itemize}

\setlength\itemsep{0.15em}

\item Challenge 1: Extremely large model size. Despite the high density of memristors, the capacity of the crossbar is still limited by peripheral circuits such as DAC (\textit{digital-analog converter}) and ADC (\textit{analog-digital converter}), which occupy a significant portion of the chip. Using traditional architectures, it is impossible to deploy the entire Large Language Model (LLM) on a single chip. For instance, the GPT-3 model has over 175 billion parameters, and it would require \textbf{2777} ISAAC chips~\cite{shafiee2016isaac} to store all of its parameters. This significantly weakens the advantages of using memristor crossbars over TPU and GPU accelerators. Today, with the continuous growth of LLMs, the capacity of traditional memristor crossbars has become a major bottleneck.

\item Challenge 2: Non-weight stationary computations. Traditional memristor crossbars are designed for weight-stationary matrix multiplication, where one of the operands is weights that can be pre-stored into the memristors. This is because dynamically programming the memristors and changing their values is both time and energy-consuming for model inference. However, for most language models that contain multi-head attention blocks, non-weight multiplication is inevitable. For example, we need to compute the matrix multiplication among the query, key, and value matrix. In these cases, both of the operands are intermediate results from the upstream operations. These non-weight stationary multiplications make it difficult to deploy LLMs directly on the memristor crossbar. 

\item Challenge 3: Complex non-linear operations.
The memristor crossbars excel primarily in performing regular linear multiplications, which are relatively straightforward computations. However, LLM architectures usually incorporate numerous nonlinear operations such as Softmax, LayerNorm, and others. These non-linear operations often require several steps to compute. For instance, we use softmax to normalize an array of elements. To achieve this, the exponential value of each element is computed, and these values are then summed before the normalization step takes place. The existence of these complex non-linear operations in LLMs makes it challenging to deploy them on memristor crossbars. 

\end{itemize}

\vspace{0.1em}

We propose a new architecture that enables energy-efficient model inference of LLM on memristor-based machine learning accelerators. This new architecture is capable of producing computation results that are highly comparable to those of traditional accelerators, with negligible accuracy loss when compared to state-of-the-art devices such as TPUs and GPUs. In summary, the proposed memristor architecture is capable of the following:

\vspace{0.1em}

\begin{itemize}

\setlength\itemsep{0.15em}

\item Fit an entire LLM on a single chip, eliminating the extra time and energy caused by off-chip communications.

\item Compatible with non-weight stationary multiplication in multi-head attention blocks of LLM.

\item Able to execute all the operations in LLM by decomposing them into standardized sub-operations.

\item Has lower energy consumption, area, and is more robust over a wide range of applications.

\end{itemize}

\vspace{0.1em}

The paper is structured as follows: Section 2 presents a literature review of related works. Section 3 provides background knowledge on the LLM and memristor crossbar. Section 4 introduces our method that can decompose all the operations in LLM into standardized sub-operations. Section 5 introduces our proposed memristor crossbar architecture that can support the execution of sub-operations. In Section 6, we quantitatively analyze our design and compare it with state-of-the-art approaches. Finally, we draw conclusions in Section 7.

\section{Related Work}

Before the emergence of non-volatile memory, two major categories of memory devices were prevalent~\cite{zahoor2020resistive}: SRAM (\textit{static random-access memory}) and DRAM (\textit{dynamic random-access memory}). An SRAM cell consists of six transistors, occupying a relatively larger area~\cite{yeap20195nm}. On the other hand, a DRAM cell comprises a transistor and a capacitor, storing one bit of information~\cite{dram}.
With the development of emerging memory systems like RRAM (resistive random access memory), even higher density has been achieved. Each RRAM cell consists of a memristor (3D-stacked) and a transistor~\cite{hu2016dot}. Unlike DRAM and SRAM, the RRAM is capable of storing multiple bits in each cell, rather than just one bit~\cite{stathopoulos2017multibit}. Utilizing RRAM increases the chance to deploy large-scale neural network models within a single chip or package, thus effectively bypassing inefficient off-chip communication.

Memristor crossbars have shown great potential in computing deep neural networks (DNNs) with higher energy efficiency than traditional neural network accelerators such as TPU and GPU, in traditional computer vision applications~\cite{hu2016dot}. There are two basic architectures using two different formats of data storage: multi-bit memristor and single-bit memristor. In the multi-bit design, each memristor stores multiple bits of information. Examples include~\cite{stathopoulos2017multibit,agrawal2020memory}. This architecture has a very high density of data storage as each weight only occupies one or a few memristors. Another design is the single-bit memristor, where we need multiple memristors to store a single weight. Examples are~\cite{zhu2019configurable,xue202116}. Since the data is computed in a digital way with binarized data, The design excels in handling unstable environments like noise, but it sacrifices density for robustness.

In the literature, several improvements have been made to the basic architectures of memristor crossbars to increase their computing efficiency on DNNs. Among them, PRIME~\cite{chi2016prime}, ISAAC~\cite{shafiee2016isaac}, and PipeLayers~\cite{song2017pipelayer} are the most popular and typical designs. These architectures combine features from both multi-bit and single-bit designs and demonstrate higher energy efficiency, throughput, and computation density compared to traditional accelerators. Another type of device that people commonly used to compute neural networks analogously today is SRAM crossbar. For example, Vesti is an SRAM-based neural network accelerator~\cite{yin2019vesti}. Compared to NVM, SRAM crossbar is easier to reconfigure the memory content. However, its density is only 4\% of traditional NVM devices such as RRAM~\cite{zahoor2020resistive}, making it unsuitable for storing extremely large neural network models.

\begin{figure}[!t]
\centering
\includegraphics[width=3.5in]{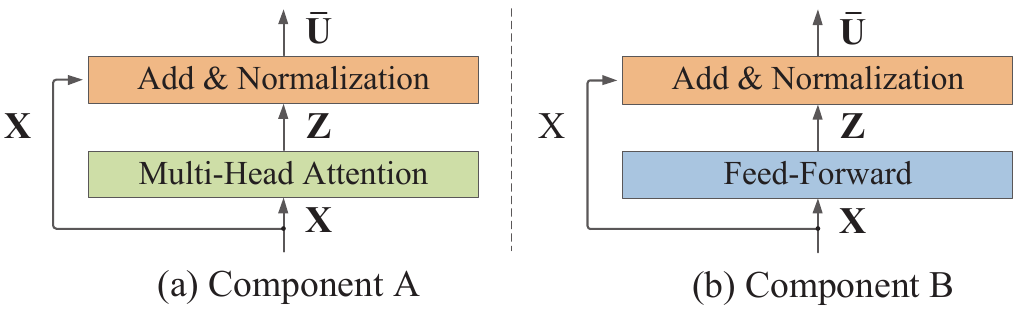}\\
\caption{Two basic components that compose layers in LLM. For example, a layer in BERT consists of one component A and one component B.}
\label{f:blocks}
\end{figure}

\section{Preliminaries}

In this section, we will first introduce the background knowledge of language models and their internal operations. Next, we will present the typical architecture of the memristor crossbar and discuss its characteristics.

\subsection{Operations in Large Language Models}

Take the language model BERT as an example. BERT utilizes a series of encoder layers to process input data. Each encoder layer in BERT consists of two fundamental components, represented as component A and component B in Figure~\ref{f:blocks}. In other language models, the layers are also structured using the same two components (Component A is masked in some models). 
Component A comprises a multi-head attention block followed by an add and normalization block. Within this component, the multi-head attention block takes the input denoted as \textbf{X} and produces an output denoted as \textbf{Z}. The add and normalization block then takes the sum of \textbf{X} and \textbf{Z} as input and generates an output denoted as $\bar{\textbf{U}}$. On the other hand, Component B consists of a feed-forward block followed by an add and normalization block. The notation for the input \textbf{X}, output \textbf{Z}, and final output $\bar{\textbf{U}}$ in Component B follow a similar convention as used in Component A.

\renewcommand{\arraystretch}{1.2}

\begin{table}[!t]
\caption{Operations in LLM. The abbreviations WS and NW stand for weight-stationary and non-weight-stationary, respectively. }
\resizebox{1.00\linewidth}{!}{
\begin{tabular}{c | l |c ? c| l|c}
\toprule
\multicolumn{6}{c}{(a) Multi-Head Attention} \\
\midrule
No. &Formulation & Type  & No. & Formulation &Type \\
\midrule
1 & $\textbf{X}\cdot \textbf{W}_q=\textbf{Q}\ \ \ \ \ \ $ &WS & 4 & $\textbf{Q}\cdot \textbf{K}^T/\sqrt{d_k}=\textbf{S}$  &NW \\ 

2 & $\textbf{X}\cdot \textbf{W}_k=\textbf{K}$ &WS & 5 & $\textrm{Softmax}(\textbf{S})\cdot \textbf{V}=\textbf{Y}$ &NW \\

3 & $\textbf{X}\cdot \textbf{W}_v=\textbf{V}$ &WS &6 & $\textbf{Y}\cdot \textbf{W}_o=\textbf{Z}$&WS  \\ 
\bottomrule
\multicolumn{6}{l}{ }  \\[-1ex]
\toprule
\multicolumn{6}{c}{(b) Feed-Forward}\\
\midrule
No. &Formulation & Type  & No. & Formulation &Type \\
\midrule
1 & $\textbf{X}\cdot \textbf{W}_a=\textbf{Y} $ &WS & 2 &  $\textrm{ReLU}(\textbf{Y})
\cdot \textbf{W}_b=\textbf{Z}$  &WS \\ 
\bottomrule
\multicolumn{6}{l}{ }  \\[-1ex]
\toprule
\multicolumn{6}{c}{(c) Add \& Normalization}\\
\midrule
No. &\multicolumn{4}{l|}{Formulation} &Type \\
\midrule
1 & \multicolumn{4}{l|}{$\bar{\textbf{U}}=\textrm{LayerNorm}(\textbf{Z}+\textbf{X})=\textrm{LayerNorm}(\textbf{U})$ } & NW\\
\bottomrule
\end{tabular}
}
\label{t:pres}
\end{table}

The details of the multi-head attention block are shown in Table~\ref{t:pres}(a). Assuming the input sequence matrix \textbf{X}, it is first multiplied with three weight matrices individually: \textbf{W}$q$, \textbf{W}$k$, and \textbf{W}$v$. This generates matrices \textbf{Q}, \textbf{K}, and \textbf{V}, respectively. The next step is to multiply the matrix \textbf{Q} with \textbf{K}$^T/\sqrt{d_k}$, which represents the transposed \textbf{K} scaled by a factor $/\sqrt{d_k}$. Here $d_k$ is a fixed value indicating the width of the matrix \textbf{K}. This multiplication operation results in a matrix \textbf{S}, representing the attention scores between the query and the key. Next, we apply the softmax function to normalize \textbf{S}, and the result is denoted as Softmax(\textbf{S}). This softmax function ensures that the values in each row of Softmax(\textbf{S}) range from 0 to 1 and sum up to 1, representing the importance of relative elements. 
It is defined in Equation~(\ref{e:softmax}).
\begin{equation}
s^{\prime}_{ij} = e^{s_{ij}}/( e^{s_{i1}}+e^{s_{i2}}+\cdots + e^{s_{in}})
\label{e:softmax}
\end{equation}
Here, $s_{ij}$ represents the element in matrix \textbf{S}, while $s^{\prime}_{ij}$ represents the corresponding element in matrix Softmax(\textbf{S}).
Once we obtained Softmax(\textbf{S}), we multiply it with matrix \textbf{V}, yielding a matrix \textbf{Y}. Subsequently, we perform a matrix multiplication between the matrix \textbf{Y} and another weight matrix \textbf{W}$_o$, producing the output matrix \textbf{Z}. 

The details of the feed-forward block are illustrated in Table~\ref{t:pres}(b). The process begins with the input matrix \textbf{X}, which undergoes multiplication by the weight matrix \textbf{W}$_a$, resulting in an intermediate matrix \textbf{Y}. Subsequently, \textbf{Y} is multiplied by another weight matrix \textbf{W}$_b$, producing the final output matrix \textbf{Z}. 

The details of the add and normalization block can be found in Table~\ref{t:pres}(c). There is one operation called LayerNorm, which transforms the matrix \textbf{U} (the sum of \textbf{X} and \textbf{Z}) into $\bar{\textbf{U}}$. This transformation is defined by Equation~(\ref{e:layernorm}). 
\begin{equation}
\bar{u}_{ij}=\frac{u_{ij}-E(u_{i\ast})}{\sqrt{\textrm{Var}(u_{i\ast})+\epsilon}}\cdot \gamma+\beta
\label{e:layernorm}
\end{equation}

In this equation, $u_{ij}$ and $\bar{u}_{ij}$ represent the corresponding elements in matrix \textbf{U} and $\bar{\textbf{U}}$, respectively. The parameters $\epsilon$, $\gamma$, and $\beta$ are trainable parameters and remain fixed during inference. The terms $E(u_{i\ast})$ and $\textrm{Var}(u_{i\ast})$ denote the mean and variance, respectively, of the elements in the $i$-th row of the matrix \textbf{U}.

As indicated in Table~\ref{t:pres}, operations $4$ and $5$ in (a), as well as operation $1$ in (c), are classified as non-weight stationary (NW) computations. On the other hand, the remaining operations are all categorized as weight-stationary (WS) computations.

\begin{figure}
\centering
\subfigure[]{\includegraphics[width=1.23 in] {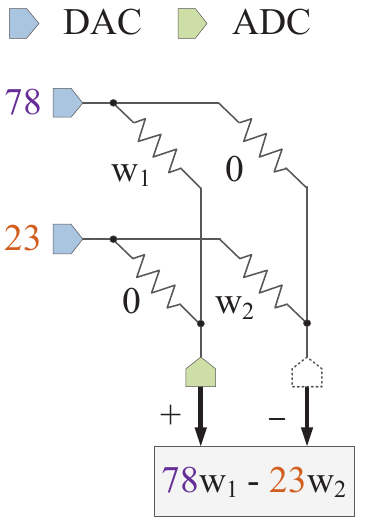}} 
\subfigure[]{\includegraphics[width=2.22 in]{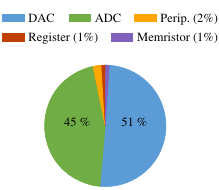}}
\caption{(a) A multi-bit memristor crossbar with $2\times 2$ cells ; (b) The area breakdown of a 128x128 memristor crossbar~\cite{shafiee2016isaac}, using a shared ADC.}
\label{fig:background}
\end{figure}

\begin{table*}[!t]
\caption{We decompose the operations within the \textbf{multi-head attention} block into standardized sub-operations, illustrated in Equation~(\ref{e:unit}) and~(\ref{e:unit2}). The third column uses the abbreviations WS and NW to represent weight-stationary and non-weight stationary, respectively.}
\resizebox{1.00\linewidth}{!}{
\begin{tabular}{c | c |c |c| c|c|c| c}
\toprule
No. &$F(\mathbb X\cdot \mathbb Y)=F(\mathbb Z)$ & Type  & $\mathbb X  \cdot\textrm{col}_t(\mathbb Y)$ & $\textrm{col}_t(\mathbb Z)$& $F$& $F(\textrm{col}_t(\mathbb Z))$ & $t\in$\\
\midrule
1 &${\textbf{XW}_q}={\textbf{Q}}$ & WS&$
\begin{bmatrix}
   x_{11}  & \cdots & x_{1m} \\
   \vdots    & \ddots & \vdots  \\
   x_{n1} & \cdots & x_{nm} 
\end{bmatrix}
$ 
 $
\begin{bmatrix}
   w^q_{1t}  \\
   \vdots   \\
   w^q_{mt}  
\end{bmatrix}
$ & $
\begin{bmatrix}
   q_{1t} = x_{11}w^q_{1t}+\cdots x_{1m}w^q_{mt}  \\
   \vdots   \\
    q_{nt} = x_{n1}w^q_{1t}+ \cdots x_{nm}w^q_{mt}  \\ 
\end{bmatrix}
$ &N.A. &$\begin{bmatrix}
   q_{1t}  \\
   \vdots   \\
   q_{nt}  
\end{bmatrix}
$ & $[1,m]$ \\ 
\midrule
2 &$\dfrac{\textbf{XW}_k}{\sqrt{d_k}}=\dfrac{\textbf{K}}{\sqrt{d_k}}$  & WS &$
\begin{bmatrix}
   x_{11}  & \cdots & x_{1m} \\
   \vdots    & \ddots & \vdots  \\
   x_{n1}  & \cdots & x_{nm} 
\end{bmatrix}
$ 
$
\begin{bmatrix}
   w^k_{1t}  \\
   \vdots   \\
   w^k_{mt}  
\end{bmatrix}
$ & $
\begin{bmatrix}
   k_{1t} = x_{11}w^k_{1t}+\cdots x_{1m}w^k_{mt}  \\
   \vdots   \\
    k_{nt} = x_{n1}w^k_{1t}+ \cdots x_{nm}w^k_{mt}  \\ 
\end{bmatrix}
$ &$ /{\sqrt{d_k}}$
&$\begin{bmatrix}
   \dfrac{k_{1t}}{\sqrt{d_k}}  \\
   \vdots   \\
   \dfrac{k_{nt}}{\sqrt{d_k}} 
\end{bmatrix}
$ & $[1,m]$ \\
\midrule
3 &$\textbf{XW}_v=\textbf{V}$  & WS&$
\begin{bmatrix}
   x_{11} & \cdots & x_{1m} \\
   \vdots  s  & \ddots & \vdots  \\
   x_{n1}  & \cdots & x_{nm} 
\end{bmatrix}
$ 
$
\begin{bmatrix}
   w^v_{1t}  \\
   \vdots   \\
   w^v_{mt}  
\end{bmatrix}
$ & $
\begin{bmatrix}
   v_{1t} = x_{11}w^v_{1t}+\cdots x_{1m}w^v_{mt}  \\
   \vdots   \\
    v_{nt} = x_{n1}w^v_{1t}+ \cdots x_{nm}w^v_{mt}  \\ 
\end{bmatrix}
$ &N.A.&$\begin{bmatrix}
   v_{1t}  \\
   \vdots   \\
   v_{nt}  
\end{bmatrix}
$ & $[1,m]$ \\
\midrule
4 &EXP$\{\dfrac{\textbf{QK}^T}{\sqrt{d_k}}\}=$ EXP$\{\textbf{S}\}$  & NW&$
\begin{bmatrix}
   q_{11} & \cdots & q_{1m} \\
   \vdots    & \ddots & \vdots  \\
   q_{n1}  & \cdots & q_{nm} 
\end{bmatrix}
$ 
 $
\begin{bmatrix}
   \dfrac{k_{t1}}{\sqrt{d_k}}  \\
   \vdots   \\
   \dfrac{k_{tm}}{\sqrt{d_k}} 
\end{bmatrix}
$ & $
\begin{bmatrix}
   s_{1t} = \dfrac{q_{11}k_{t1}}{\sqrt{d_k}}+ \cdots \dfrac{q_{1m}k_{tm}}{\sqrt{d_k}}  \\
   \vdots   \\
   s_{nt} = \dfrac{q_{n1}k_{t1}}{\sqrt{d_k}}+ \cdots \dfrac{q_{nm}k_{tm}}{\sqrt{d_k}}  
\end{bmatrix}
$ &EXP&$\begin{bmatrix}
    e^{s_{1t}}  \\
   \vdots   \\
    e^{s_{nt}}  
\end{bmatrix}
$ & $[1,n]$ \\ 
\midrule
5 &EXP$\{\textbf{S}\}\cdot \textbf{1}=\textbf{a}$  & NW&$
\begin{bmatrix}
   e^{s_{11}}  & \cdots & e^{s_{1n}} \\
   \vdots    & \ddots & \vdots  \\
   e^{s_{n1}}  & \cdots & e^{s_{nn}} \\
\end{bmatrix}
$ 
$
\begin{bmatrix}
   1  \\
   \vdots   \\
  1  
\end{bmatrix}
$ & $
\begin{bmatrix}
   a_{1} = e^{s_{11}}+e^{s_{12}}+ \cdots e^{s_{1n}}  \\
\vdots   \\
   a_{n} = e^{s_{n1}}+e^{s_{n2}}+ \cdots e^{s_{nn}}
\end{bmatrix}
$ &N.A.&$\begin{bmatrix}
   a_{1}  \\
   \vdots   \\
   a_{n}  
\end{bmatrix}
$ & \{1\} \\
\midrule
6 &{
\begin{tabular}{@{}c@{}}EXP$\{\textbf{S}\}\textbf{V}\oslash (\textbf{a}\cdot \textbf{1}^T)$  \\ $=\textbf{R}\oslash (\textbf{a}\cdot \textbf{1}^T)$\end{tabular}}
  & NW&{$
\begin{bmatrix}
   e^{s_{11}}  & \cdots & e^{s_{1n}} \\
   \vdots    & \ddots & \vdots  \\
   e^{s_{n1}}  & \cdots & e^{s_{nn}} \\
\end{bmatrix}
$ 
 $
\begin{bmatrix}
  {v_{1t}}  \\
   \vdots   \\
  {v_{nt}}  
\end{bmatrix}
$} & {$ 
\begin{bmatrix}
   r_{1t} =  e^{s_{11}}v_{1t} +\cdots e^{s_{1n}}v_{nt}   \\
   \vdots   \\
    r_{nt} =  e^{s_{n1}}v_{1t}+  \cdots  e^{s_{nn}}v_{nt}  \\ 
\end{bmatrix}
$ }&{ $/{a_i}$ } &{$\begin{bmatrix}
   \dfrac{r_{1t}}{a_1}  \\
   \vdots   \\
   \dfrac{r_{nt}}{a_n}  
\end{bmatrix}
$ }& {$[1,m]$} \\ 
\midrule
7 &
\begin{tabular}{@{}c@{}}${\textbf{R}\oslash (\textbf{a}\cdot \textbf{1}^T)\cdot \textbf{W}_o}+{\textbf{X}}$  \\ $=\textbf{Z}+\textbf{X}$\end{tabular}
   & WS &$
\begin{bmatrix}
   \dfrac{r_{11}}{a_1}  & \cdots & \dfrac{r_{1m}}{a_1} \\
   \vdots    & \ddots & \vdots  \\
   \dfrac{r_{n1}}{a_n}  & \cdots & \dfrac{r_{nm}}{a_n} \\
\end{bmatrix}
$ 
 $
\begin{bmatrix}
   w^o_{1t}  \\
   \vdots   \\
   w^o_{mt}  
\end{bmatrix}
$ & $
\begin{bmatrix}
   z_{1t} = \dfrac{r_{11}}{a_1}w^o_{1t}+ \cdots \dfrac{r_{1m}}{a_1}w^o_{mt}  \\
   \vdots   \\
    z_{nt} = \dfrac{r_{n1}}{a_n}w^o_{1t}+  \cdots \dfrac{r_{nm}}{a_n}w^o_{mt}  \\ 
\end{bmatrix}
$ &$+x_{it}$ &$\begin{bmatrix}
   z_{1t}+x_{1t}  \\
   \vdots   \\
   z_{nt}+x_{nt} 
\end{bmatrix}
$ & $[1,m]$ \\ 
\bottomrule
\end{tabular}
}
\label{t:mh}
\end{table*}

\subsection{Traditional Memristor Crossbar}

Traditional memristor-based machine learning accelerators are optimized for weight-stationary matrix multiplications, making them well-suited for deploying most computer vision (CV) models. Figure~\ref{fig:background}(a) provides an example of a memristor crossbar with $2\times 2$ cells. This is a multi-bit design. Each cell stores a single parameter of the neural network, with the conductance of the cell representing the weight stored. A \textit{digital-to-analog converter} (DAC) is used to transform activation values into voltages applied to the memory cell. By Ohm's Law, the current flowing through the cell equals the applied voltage multiplied by the conductance. Finally, an \textit{analog-to-digital converter} (ADC) converts the combined current from various memory cells back into digital data. Kirchhoff's Current Law ensures that the current sum is equivalent to the summation of the products.

Despite the higher density of memristors compared to DRAM and SRAM, the effective density of the entire system remains low due to the necessity of implementing high-bit DACs/ADCs in the input and output circuits of the crossbar, which occupy a significant area on the chip. The area breakdown of a classical $128\times128$ memristor crossbar in Figure~\ref{fig:background} (b) illustrates this. Even with the sharing of the ADC among the 128 columns~\cite{shafiee2016isaac}, the DACs and ADCs still demand approximately 51\% and 45\% of the total area, respectively. The ISAAC~\cite{shafiee2016isaac} crossbar architecture reduces the area overhead of the DAC at the expense of longer computation time. Despite this, only around 2\% of the area being allocated to memristors. While the PRIME~\cite{chi2016prime} and PipeLayer~\cite{song2017pipelayer} approaches eliminate the need for ADCs, their input or output circuits still contain numerous capacitors, which consume a significant amount of chip area.

The substantial area overhead caused by implementing input and output circuits (e.g., DACs and ADCs) within the memristor crossbar, as noted in previous studies~\cite{saberi2011analysis,kull20133}, diminishes the advantage of the high density offered by RRAM devices in comparison to alternative techniques. Due to the considerable time and energy consumption associated with memristor programming, dynamic modification of stored values during inference is infrequent. It becomes necessary to pre-store the entire set of parameters within the memristors. However, because of the substantial area overhead of peripheral circuits, the total area overhead for the large neural network model is huge. Therefore, with classical architecture design for memristor crossbar, we have to deploy the large model on multiple chips with inefficient off-chip communication in terms of time and energy. This significantly limits the application of the memristor crossbar in LLM.

\begin{table*}[!t]
\caption{We decompose the operations within the \textbf{feed-forward} block into standardized sub-operations, illustrated in Equation~(\ref{e:unit}) and~(\ref{e:unit2}). The third column uses the abbreviations WS and NW to represent weight-stationary and non-weight stationary, respectively.}
\resizebox{1.00\linewidth}{!}{
\begin{tabular}{c | c |c |c| c|c|c| c}
\toprule
No. &$F(\mathbb X\cdot \mathbb Y)=F(\mathbb Z)$ & Type  & $\mathbb X  \cdot\textrm{col}_t(\mathbb Y)$ & $\textrm{col}_t(\mathbb Z)$& $F$& $F(\textrm{col}_t(\mathbb Z))$ & $t\in$\\
\midrule
1 &\begin{tabular}{@{}c@{}}ReLU(${\textbf{XW}_a}+\textbf{b}_a)$   \\ $=$ ReLU$(\textbf{Y})$\end{tabular}&WS &$
\begin{bmatrix}
   x_{11} & \cdots   & x_{1m} & 1 \\
   \vdots & \ddots   & \vdots  & \vdots  \\
   x_{n1} & \cdots  & x_{nm} & 1 
\end{bmatrix}
$ 
 $
\begin{bmatrix}
   w^a_{1t}  \\
   \vdots   \\
   w^a_{mt}  \\ 
   b^a_{t} 
\end{bmatrix}
$ & $
\begin{bmatrix}
   y_{1t} = x_{11}w^a_{1t}+\cdots x_{1m}w^a_{mt}+ b^a_{t} \\
   \vdots   \\
    y_{nt} = x_{n1}w^a_{1t}+ \cdots x_{nm}w^a_{mt} + b^a_{t}\\ 
\end{bmatrix}
$ & ReLU &$\begin{bmatrix}
   \textrm{ReLU}(y_{1t})  \\
   \vdots   \\
   \textrm{ReLU}(y_{nt})  
\end{bmatrix}
$ & $[1,h]$ \\ 
\midrule
2 &\begin{tabular}{@{}c@{}}ReLU$(\textbf{Y)}\cdot\textbf{W}_b+\textbf{b}_b$  \\ $+\ \textbf{X}=\textbf{Z}+\textbf{X}$ \end{tabular}   & WS &$
\begin{bmatrix}
   y_{11}  & \cdots & y_{1h} & 1 \\
   \vdots   & \ddots & \vdots  & \vdots  \\
   y_{n1}  & \cdots & y_{nh} & 1 
\end{bmatrix}
$ 
$
\begin{bmatrix}
   w^b_{1t}  \\
   \vdots   \\
   w^b_{ht}  \\ 
   b^b_{t} 
\end{bmatrix}
$ & $
\begin{bmatrix}
   z_{1t} = y_{11}w^b_{1t}+\cdots y_{1h}w^b_{ht}+ b^b_{t} \\
   \vdots   \\
    z_{nt} = y_{n1}w^b_{1t}+ \cdots y_{nh}w^b_{ht} + b^b_{t}\\ 
\end{bmatrix}
$ & $+x_{it}$
&$\begin{bmatrix}
   z_{1t}+x_{1t}   \\
   \vdots   \\
   z_{nt}+x_{nt}  
\end{bmatrix}
$ & $[1,m]$ \\
\bottomrule
\end{tabular}
}
\label{t:ff}
\end{table*}

\begin{table*}[!t]
\caption{We decompose the operations within the \textbf{layer normalization} function into standardized sub-operations, illustrated in Equation~(\ref{e:unit}) and~(\ref{e:unit2}). The third column uses the abbreviations WS and NW to represent weight-stationary and non-weight stationary, respectively.}
\resizebox{1.00\linewidth}{!}{
\begin{tabular}{c | c |c |c| c|c|c| c}
\toprule
No. &$F(\mathbb X\cdot \mathbb Y)=F(\mathbb Z)$ & Type  & $\mathbb X  \cdot\textrm{col}_t(\mathbb Y)$ & $\textrm{col}_t(\mathbb Z)$& $F$& $F(\textrm{col}_t(\mathbb Z))$ & $t\in$\\
\midrule
1 & $E(u_{t\ast})=\ $row$_t(\textbf{U})\cdot \textbf{1}$ &NW &$
\begin{bmatrix}
   u_{t1} & u_{t2}  & \cdots & u_{tm}  \\
\end{bmatrix}
$ 
 $
\begin{bmatrix}
    1  \\
   \vdots   \\
    1  \\
\end{bmatrix}
$ & $
  \mathlarger{\sum}_{i=1}^{m} u_{ti}= u_{t1}+  u_{t2}+\cdots+  u_{tm}
$  &\ \ \  /m\ \ \   &$\mathlarger{\sum}_{i=1}^{m} u_{ti}/m$  & $[1,n]$ \\ 
\midrule
2 & $E(u^2_{t\ast})=\ $row$_t(\textbf{U})\cdot $row$_t(\textbf{U})^T$ &NW &$
\begin{bmatrix}
   u_{t1} & u_{t2}  & \cdots & u_{tm}  \\
\end{bmatrix}
$ 
 $
\begin{bmatrix}
    u_{t1}  \\
   \vdots   \\
    u_{tm}  \\
\end{bmatrix}
$ & $
  \mathlarger{\sum}_{i=1}^{m} u^2_{ti}= u^2_{t1}+  u^2_{t2}+\cdots+  u^2_{tm}
$  &/m &$\mathlarger{\sum}_{i=1}^{m} u^2_{ti}/m$  & $[1,n]$ \\ 
\bottomrule
\end{tabular}
}
\label{t:norm}
\end{table*}

\begin{figure}[t]
\centering
\includegraphics[width=3.5in]{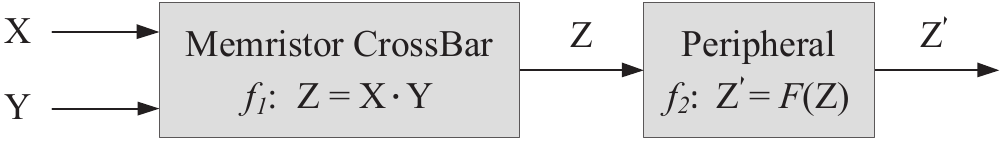}\\
\caption{{The sub-operation has two phases. The first phase involves the linear multiplication between \textbf{X} and \textbf{Y}, while the second phase involves the addition function $F$ applied to the multiplication result \textbf{Z}.}}
\label{f:phasesub}
\end{figure}

\section{Standardized Operation Decomposition}

The LLM consists of a wide range of operations, including both linear operations such as weight stationery and non-weight stationary multiplication, as well as non-linear operations like softmax and layer normalization. The diversity of operations within the LLM presents challenges for a single hardware module to efficiently perform all the required functions. Without optimization, it would be necessary to employ separate hardware modules to handle the diverse computational requirements. To enable the seamless implementation of LLM on a unified hardware module, we decompose all the operations of LLM into standardized sub-operations, as illustrated in Equation~(\ref{e:unit}).
\begin{equation}
F(\mathbb X\cdot \mathbb Y)=F(\mathbb Z)
\label{e:unit}
\end{equation}
{As depicted in Figure~\ref{f:phasesub}, each standardized sub-operation within the LLM consists of a fundamental linear operation executed by memristor-based crossbars and an additional $F$ executed by peripheral module.} The linear operation involves multiplying a matrix $\mathbb X$ with a matrix $\mathbb Y$, resulting in a matrix $\mathbb Z$. Depending on the specific context, this linear operation can be either weight stationary or non-weight stationary. In the case of weight-stationary multiplication, the matrix $\mathbb Y$ can be replaced by a weight matrix denoted as $\mathbb W$. The additional operator $F$ can be either linear, such as multiplication, addition, or non-linear, incorporating functions like the exponential function (EXP), Rectified Linear Unit (ReLU), division, and various others.

For easier hardware implementation, we can further decompose each sub-operation into multiple sessions, as shown in Equation~(\ref{e:unit2}). 
We denote $\textrm{col}_t(\mathbb Y)$ and $\textrm{col}_t(\mathbb Z)$ as the $t$-th column of matrices $\mathbb Y$ and $\mathbb Z$, respectively. In each session, we only compute the multiplication between $\mathbb X$ and the vector col$_t(\mathbb Y)$. This operation can be performed by any hardware that supports matrix-vector multiplications.
\begin{equation}
F(\mathbb X\cdot \textrm{col}_t(\mathbb Y)) =F(\textrm{col}_t(\mathbb Z)), \ \ \   t=1,2,3\ \cdots
\label{e:unit2}
\end{equation}

The standardized operation decomposition for LLM is listed in Table~\ref{t:mh} for the multi-head attention block, Table~\ref{t:ff} for the feed-forward block, and Table~\ref{t:norm} for the layer normalization function. 
In the third column of each table, we label the type of each sub-operation. The abbreviation {WS} represents weight-stationary multiplication, and {NW} represents non-weight stationary multiplication. In the fourth and fifth columns, we provide the details of matrix $\mathbb Z$, vector $\textrm{col}_t(\mathbb Y)$, and vector $\textrm{col}_t(\mathbb Z)$. We assume the input sequence consists of $n$ tokens and the model has a hidden size of $m$. 
The sixth and seventh columns list the additional operation $F$, and the corresponding result under this function, i.e., $F(\textrm{col}_t(\mathbb Z))$. The abbreviation N.A. indicates that no additional function is required for this sub-operation. In the eighth column, we specify the range of session index $t$. For either the multi-head attention block (Table~\ref{t:mh}) and the feed-forward block (Table~\ref{t:ff}), the final output is obtained as the sum of the input \textbf{X} and the intermediate output \textbf{Z}. This sum, represented as $\textbf{Z}+\textbf{X}=\textbf{U}$, serves as the input to the subsequent add and normalization block.

\subsection{Softmax Operation}

The softmax operation is a crucial component in multi-head attention blocks. After multiplying the matrix \textbf{Q} with \textbf{K}$^T/\sqrt{d_k}$, the resulting matrix \textbf{S} undergoes a softmax operation along each row. Subsequently, the softmax result is multiplied by the matrix \textbf{V}.
The aforementioned computation process can be divided into three standardized sub-operations, represented as sub-operations $4$, $5$, and $6$ in Table~\ref{t:mh}. All of these sub-operations are non-weight stationary. The details of the decomposition process can be summarized in the following steps.

\begin{enumerate}

\item In sub-operation $4$, we multiply matrix $\textbf{Q}$ with $\textbf{K}^T/\sqrt{d_k}$ to obtain matrix $\textbf{S}$. This operation is performed in $n$ sessions. During session $t$, matrix $\textbf{Q}$ is multiplied with the $t$-th column of matrix $\textbf{K}^T/\sqrt{d_k}$. We incorporate the additional function $F$ as EXP, which transforms each element in matrix $\textbf{S}$ from $s_{ij}$ to $e^{s_{ij}}$. As a result, we can obtain the matrix EXP$(\textbf{S})$ from this sub-operation.

\item In sub-operation $5$, we multiply the matrix EXP$(\textbf{S})$ with vector $\textbf{1}$, where vector $\textbf{1}$ is defined as a vector with all elements being $1$. This sub-operation is performed in a single session $(t\in\{1\})$, with no additional operation $F$ involved. Each element in the final output vector, denoted by $a_t$, is the summation of all the elements in the $i$-th row of matrix EXP$(\textbf{S})$, as illustrated in Equation~(\ref{e:soft}).

\begin{equation}
a_i = e^{s_{i1}}+e^{s_{i2}}+\cdots + e^{s_{in}} = \sum_j e^{s_{ij}}
\label{e:soft}
\end{equation}

\item {In sub-operation $6$, we multiply matrix EXP$(\textbf{S})$ with the matrix $\textbf{V}$, resulting in matrix $\textbf{R}$. This operation is also performed in $m$ sessions. During session $t$, matrix EXP$(\textbf{S})$ is multiplied with the $t$-th column of matrix $\textbf{V}$. In the output matrix, element $r_{ik}$ in matrix \textbf{R} is calculated as $\Sigma_j (e^{s_{ij}}\cdot v_{jk})$.}

\item  {Finally, we perform scaling on the matrix $\textbf{R}$. This is done by applying division as the additional operation $F$ in sub-operation $6$. Specifically, the $i$-th column of matrix $\textbf{R}$ is scaled by $a_i$. Mathematically, matrix $\textbf{R}$ is element-wise divided (symbol $\oslash$) by $( \textbf{a}\cdot \textbf{1}^T)$. This results in matrix \textbf{Y} (Table~\ref{t:pres}(a)), whose element $y_{ik}$ becomes $\Sigma_j (e^{s_{ij}}\cdot v_{jk})/{\Sigma_j e^{s_{ij}}}$, corresponding to the \textit{softmax} results.}

\end{enumerate}

\begin{figure}[t]
\centering
\includegraphics[width=3.5in]{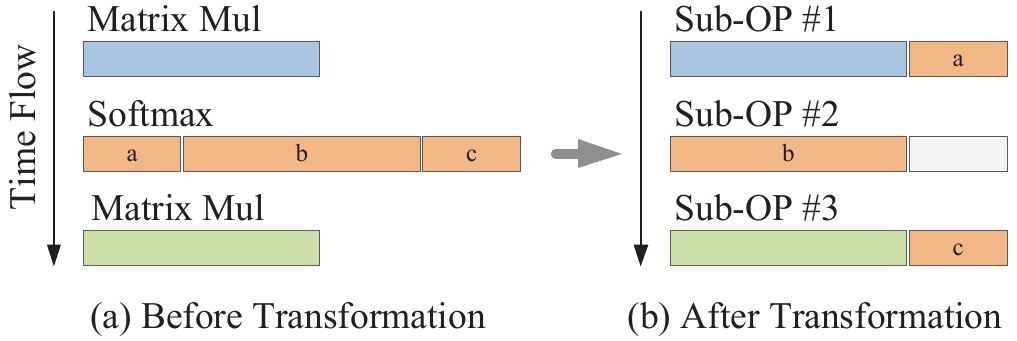}\\
\caption{{We decompose the softmax into three parts, and two of these parts can be integrated with the preceding and subsequent matrix multiplication operations, resulting in three standardized sub-operations.}}
\label{f:transfomsub}
\end{figure}

\noindent {Figure~\ref{f:transfomsub} outlines the transformation process for the softmax operation, enabling its execution on memristor-based crossbars. We begin by decomposing the softmax into three components. Afterwards, two of these components can be integrated with the matrix multiplication operations that precede and follow them, yielding three standardized sub-operations.}

\begin{figure}[!t]
\centering
\includegraphics[width=3.5in]{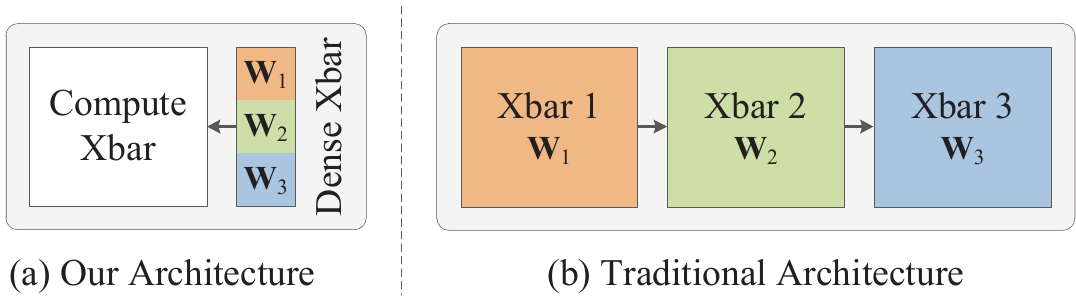}\\
\caption{Overview of our proposed architecture and the traditional architecture. We use unique colors to indicate the locations of weight matrices, denoted as $\textbf{W}_1$, $\textbf{W}_2$, and $\textbf{W}_3$. The dark arrows among the crossbars indicate the data flow direction.}
\label{f:overview1}
\end{figure}

\subsection{Layer Normalization}

Layer normalization is essential for both multi-head attention blocks and feed-forward blocks. Let's assume the input to the normalization block is $\textbf{U}=\textbf{Z}+\textbf{X}$. The normalization process is applied to each token individually, where the $t$-th token is represented by row$_t(\textbf{U})$ (i.e., the $t$-th row of the matrix \textbf{U}). We use $u_{t\ast}$ to denote elements in row$_t(\textbf{U})$. Prior to normalization, it is necessary to determine the means and variances of these elements, denoted as $E(u_{t\ast})$ and $E(u^2_{t\ast})$, respectively. These values enable us to compute the variance using Equation~(\ref{e:var}). 

\begin{equation}
\textrm{Var}(u_{t\ast}) = E(u^2_{t\ast}) - (E(u_{t\ast}))^2
\label{e:var}
\end{equation}
The calculation of the means of $u_{t\ast}$ and $u^2_{t\ast}$ can be decomposed into two standardized sub-operations, as shown in Table~\ref{t:norm}. Both of them are non-weight stationary. To calculate the sum of elements $u_{t\ast}$ in row$_t(\textbf{U})$, we can multiply this row vector with a vector $\textbf{1}$ that contains all elements as $1$. Similarly, by multiplying row$_t(\textbf{U})$ with itself, we can obtain the sum of squared elements $u^2_{t\ast}$. In both sub-operations, we incorporate additional operations $F$ multiplied by $1/m$, where $m$ represents the number of elements in row$_t(\textbf{U})$. Consequently, we can obtain $E(u_{t\ast})$ and $E(u^2_{t\ast})$ from the sub-operations.
Since there are $n$ tokens in the sequence, the above computations are executed in $n$ sessions, where each session corresponds to processing one token. Specifically, the $t$-th token in the input sequence is processed during session $t$.

After obtaining $E(u_{t\ast})$ and $E(u^2_{t\ast})$, we can utilize equation (\ref{e:var}) to calculate Var$(u_{t\ast})$. To normalize the vector, we begin by subtracting each element in the vector by its mean $E(u_{t\ast})$ and then multiply it by a scaling factor, denoted as $a$ in Equation~(\ref{e:alpha}). 
\begin{equation}
\alpha = \gamma/\sqrt{\textrm{Var}(u_{t\ast})+\epsilon}
\label{e:alpha}
\end{equation}
Here, $\gamma$ and $\epsilon$ are known parameters from the model. A dedicated module can be utilized to perform the square root and division operations. Due to the need for computation of the aforementioned process only $n$ times within each LLM layer, the requirement for such modules is minimal, resulting in a negligible area overhead.
Once we obtain $\alpha$, we have the option of directly multiplying it with $u_{t\ast}$. Alternatively, we can incorporate this multiplication as an operation $F$ within the subsequent computation blocks.

\begin{figure}[!t]
\centering
\includegraphics[width=3.5in]{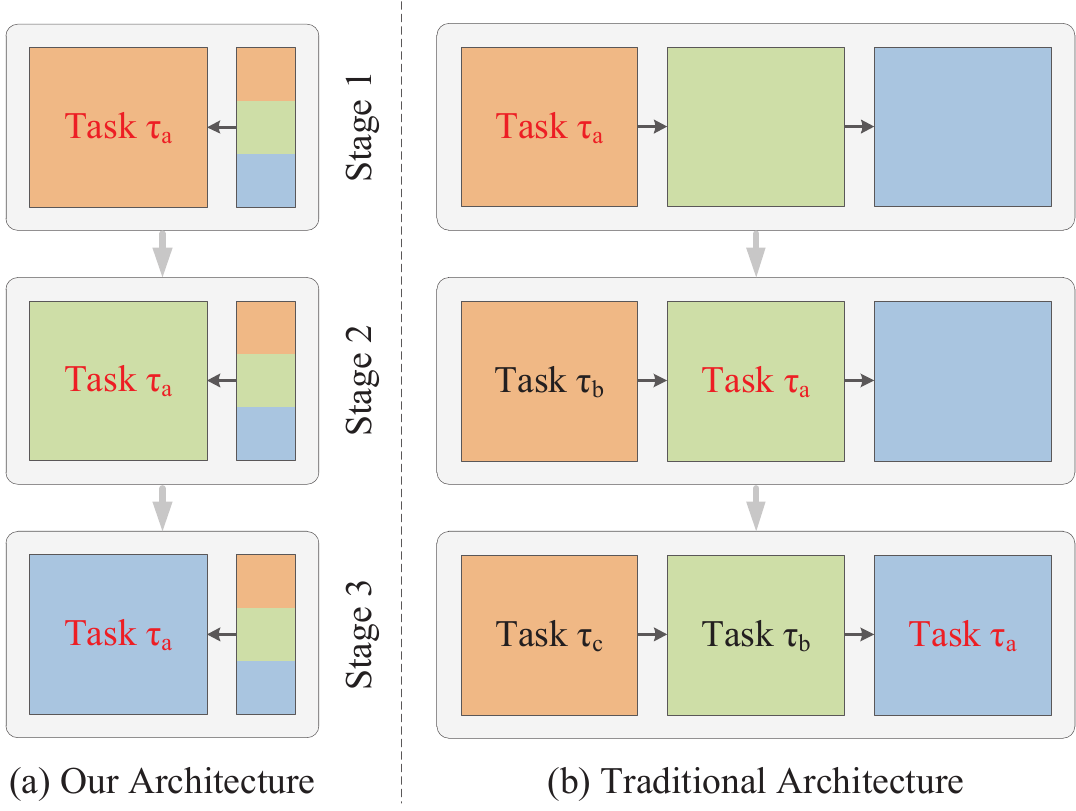}\\
\caption{A breakdown of the computation process for both architectures across multiple steps. The gray arrows indicate the time flow.}
\label{f:overview2}
\end{figure}

\section{Memristor Crossbar Architecture for Standardized Sub-operations in LLM}

We have developed an advanced architecture for memristor crossbars that enables efficient computation of the standardized sub-operations (Equation~(\ref{e:unit2})) in LLM. Our proposed system utilizes two types of crossbars: the computation crossbar, which is optimized for low-energy computing, and the dense crossbar, which is designed specifically for deploying large-scale neural networks. Both crossbar types are integrated onto the same chip to eliminate the need for inefficient off-chip communication.

\subsection{Architecture Overview}

Figure~\ref{f:overview1}(a) illustrates the overview of our proposed architecture, while Figure~\ref{f:overview1}(b) shows an example of a traditional architecture with the same weight capacity. To simplify the analysis, we assume that the model contains only three weight matrices. In both figures, we use unique colors to indicate the locations of weight matrices, denoted as $\textbf{W}_1$, $\textbf{W}_2$, and $\textbf{W}_3$. In traditional architecture (Figure~\ref{f:overview1}(b)), the weights stored within the crossbars are fixed. Weight matrices $\textbf{W}_1$ to $\textbf{W}_3$ are stored in Crossbar-$1$ to Crossbar-$3$, respectively. The dark arrows among the crossbars indicate the data flow direction. In contrast, our architecture (Figure~\ref{f:overview1}(a)) consists of computation crossbars and dense crossbars. The computation crossbar executes both weight-stationary and non-weight-stationary multiplications, where weight matrices $\textbf{W}_1$ to $\textbf{W}_3$ are stored in the dense crossbar. This results in a much smaller area overhead than the traditional architecture, due to the high area efficiency of the dense crossbars.


In Figure~\ref{f:overview2}, we present a breakdown of the computation process for both architectures across multiple stages. At the initial state, the computation crossbar is empty and contains no data information. To address this, we have developed a mechanism to instantly reconfigure the memory storage in the computation crossbar. For weight-stationary computation, the weight matrix $\textbf{W}_i$ is transferred into the computation crossbar at Stage $i$. In total, three stages are required to execute all the computations for task $\tau_a$. For non-weight stationary multiplication, another operand \textbf{Y} needs to be transferred into the compute crossbar. 

\begin{figure}[!t]
\centering
\includegraphics[width=3.5in]{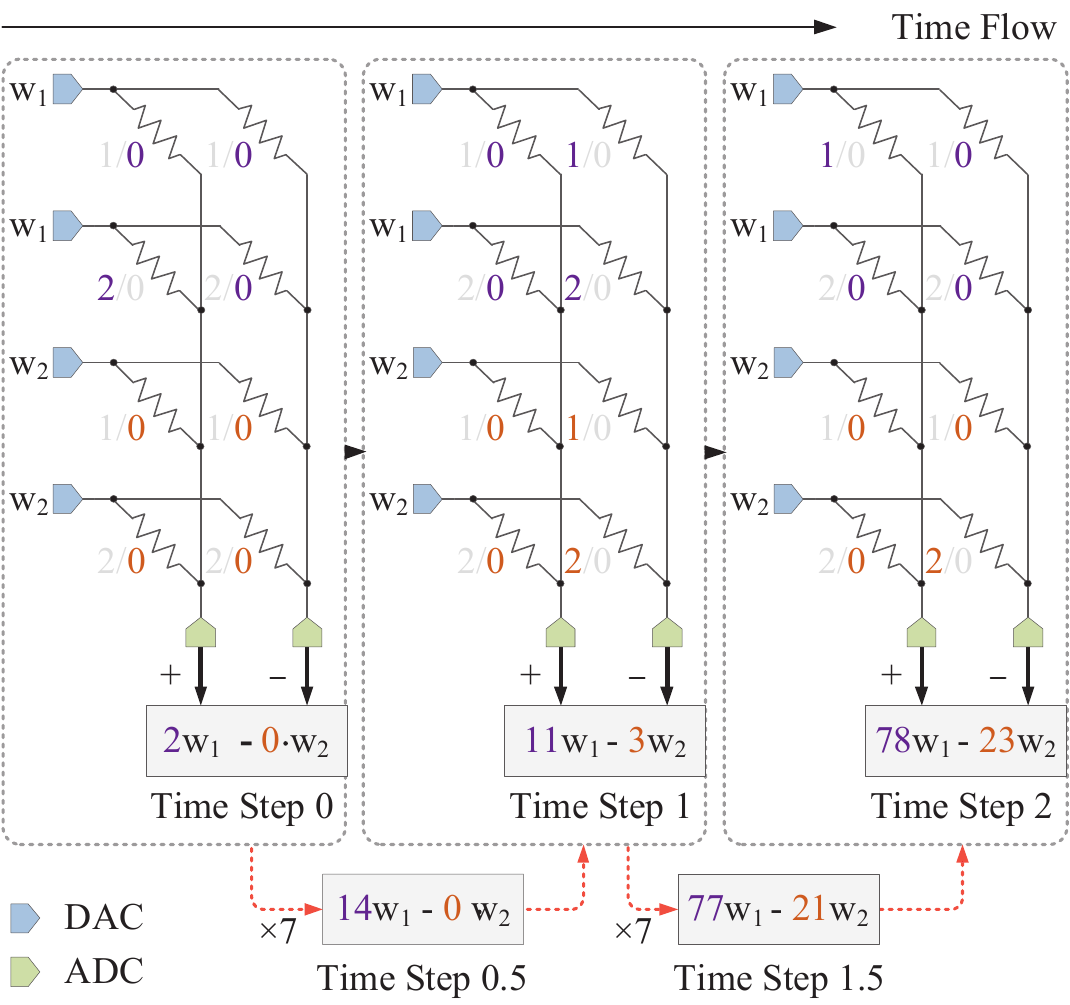}\\
\caption{Each memristor can be turned on to output fixed data value or turned off to output $0$. By controlling the on and off state of the memristor, we can perform either weight stationary or non-weight stationary multiplication in the crossbar.}
\label{f:logic}
\end{figure}

\subsection{Efficient Encoding for the Sub-Operation}

We have developed a mechanism that achieves instantaneous reconfigurability in the memristor crossbar. It is illustrated in Figure~\ref{f:logic}. In this approach, each memristor can exist in either an ``on" or an ``off" state. When in the "on" state, we can read a fixed data value, denoted as $\chi$, from the memristor. It is important to note that the value of $\chi$ remains unchanged throughout the entire computation process. On the other hand, when in the ``off" state, the memristor can only be read as $0$. To perform any multiply-accumulate (MAC) operation, we employ an encoding technique where the weights serve as input, while the activations control the state of the memristors. The activation data needs to be encoded into multiple digits to enable digit-by-digit computation. In our example, we utilize the balanced septenary (base-7) numeral system for this encoding.

In the balanced septenary numeral system, each digit can take one of seven possible values: \textbf{-3}, \textbf{-2}, \textbf{-1}, \textbf{0}, \textbf{1}, \textbf{2}, or \textbf{3}. For instance, a given data value, let's say $78$, can be encoded into three digits: \textbf{2}, \textbf{-3}, \textbf{1}. The expansion of $78$ in the balanced septenary numeral system can be expressed as $2\times 7^2 - 3\times 7^1 + 1\times 7^0 = 78$. By employing this type of encoding scheme, we can perform MAC operations efficiently, optimizing area overhead and energy consumption within the memristor crossbars. To cover all possible values of a single digit, we implement four memristors for each activation. These memristors are arranged as follows: two memristors in the positive (left) column, storing $1$ and $2$, and two in the negative (right) column, also storing $-1$ and $-2$.

To illustrate the multiplication of a positive weight $w_1$ with the activation $78$, we first obtain the three digits representing the value $78$: \textbf{2}, \textbf{-3}, \textbf{1}. Each digit corresponds to a specific computing time step. In Figure~\ref{f:logic}, the top four memristors (colored purple) represent the activation $78$.
During step-0 of computation, only the memristors storing $2$ are turned on, representing the first digit \textbf{2}. The remaining memristors are turned off and indicating value $0$. During step-1, the memristors storing $-1$ and $-2$ are activated, representing the second digit \textbf{-3}. The other memristors are turned off. Step-2 follows a similar rule, with the corresponding memristors being activated based on the third digit \textbf{1}. Between every two steps, we multiply the accumulated result by the base value $7$ since the previously processed digit holds higher significance in the overall value. If $w_1$ is negative, we exchange the states of the memristors in the two columns, reflecting the sign change. By following this process, we can effectively perform multiplication operations between the weight and the activation values, utilizing the four memristors per activation to cover all possible digit values.

\begin{figure*}[!t]
\centering
\includegraphics[width=7.15in]{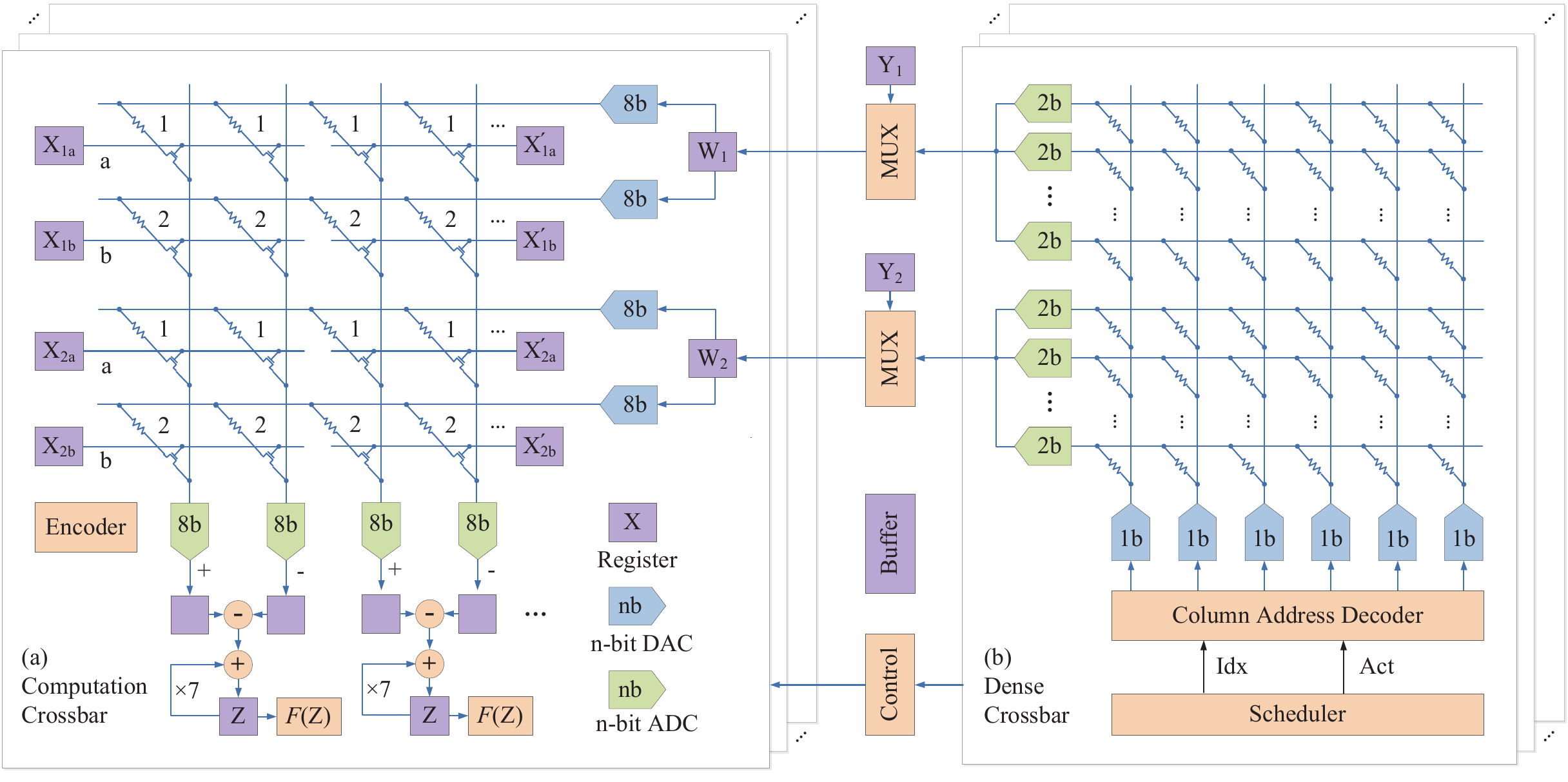}\\
\caption{(a) In the computation crossbar, we fix the data stored in each resistor and use the attached switch to control the stored data to be either $\chi$ or $0$; We have two types of rows (``a", ``b") and two types of columns (``+",``-"); (b) The dense crossbar stores all the weights of the models.}
\label{f:whole}
\end{figure*}

\subsection{Robust Computation Crossbar}

We introduce a computation crossbar that is compatible with the above encoding scheme. In this design, the memristors within the computation crossbar always store the same data value, regardless of the values of the activations. Once we establish the encoding format, there is no need to update the data stored in the memristors. This holds true even when we change neural network models. Consequently, there is no requirement to implement actual memristors and program them prior to usage. Instead, we employ regular resistors with fixed resistance to function as memristors, effectively storing the specified data. This approach enhances the resilience of the computation crossbar against random telegraph noise (RTN)~\cite{raghavan2013rtn}. Unlike memristor, the likelihood of defects occurring in the resistors is relatively low after undergoing post-fabrication examination~\cite{kayanalysis}. Consequently, the resistors do not need to operate in the low-resistance mode~\cite{ielmini2010resistance} to counteract RTN, resulting in energy savings~\cite{wang2020ncpower}.

We have devised the computation crossbar that builds upon the classic 1T1R (one-transistor-one-resistor) design~\cite{chen2015compact}. The structure of our computation crossbar can be seen in Figure~\ref{f:whole}(a). In addition to substituting memristors with conventional resistors, we have made four significant modifications to our design. These modifications are as follows:

\vspace{0.1em}

\begin{enumerate}

\setlength\itemsep{0.15em}

\item In the classical 1T1R structure~\cite{chen2015compact}, each memristor is connected in series with a transistor, serving as a control switch to regulate the current flow through the memristor. Typically, developers use this switch to enable or disable the programming functionality of the memristor during the computation phase~\cite{chen2015compact}. In our new architecture, we retain the transistor-switch design, which is attached to the memristor/resistor, and utilize the transistor to control the on-and-off state of the memristor/resistor.

\item In the balanced septenary (base-7) system, we utilize four resistors to handle each input data. These four resistors are placed into two rows (``a", ``b") and two columns (``+",``-"). To optimize the switch control, we introduce dedicated registers that are directly connected to the memristor switches, storing the control information. 
The register responsible for storing the encoded data only requires three binary bits. One bit is used to select the column, while the remaining two bits control the resistors within the selected column. The resistors in the unselected column are effectively cut off or deactivated.

\item In our computation crossbar design, we decompose the computation into multiple time steps. At each time step, the output needs to be multiplied by the base value of the encoding scheme before proceeding to the next time step. For instance, in the balanced septenary encoding system, the base value is $7$. To optimize the processing time, we can perform the multiplication by $7$ in two steps. First, we utilize shifting operations to the original value by three bits, which effectively multiplies the output by $8$. Then, we subtract the original value from this result to obtain the final multiplication by $7$. Other base values can use similar rules to optimize because all of them can be expressed as $2^{i}-1$. {These operations are executed within shift-and-add (S+A) units, whose energy consumption is relatively small compared to other components in the system~\cite{shafiee2016isaac}}

\item An module for processing additional function $F$ is implemented at the end of the linear computation. 
This function can be implemented using either a digital circuit or an analog circuit. The analog circuit also offers well-established solutions for basic operations such as exponential, multiplication, summation, and more~\cite{MacLennan2009}\cite{yawale2021operational}\cite{ulmann2013analog}. { Similar to the ADC, the peripheral module for function $F$ is shared among the columns in the crossbar in an interleaved manner~\cite{shafiee2016isaac}. Hence, its overall impact on the area cost is minimal.}

\end{enumerate}

\vspace{0.1em}

\noindent An encoder is employed to convert the activation from the original binary system into the new encoding system. Our design supports any type of balanced numeral system for encoding. In general, each activation can be encoded using $2S$ resistors within a balanced base $2^{S+1}-1$ system. The scaling factor, represented by $S$, is a crucial parameter that influences the characteristics of the encoding. We should choose the right value of $S$ based on the required precision of the activation. In this example, $S=2$ is utilized, representing the balanced septenary (base-7) encoding system. In Section 6 of our study, we will conduct a comparative analysis of different encoding schemes, ranging from $S=1$ to $7$, in order to identify the ideal encoding base value while considering a specific precision requirement for activation data.

\subsection {Dense Crossbar with High Capacity}

Our system is specifically designed to support large-scale neural network models by utilizing an additional crossbar with a substantial storage capacity. This new hardware is referred to as the ``dense crossbar" due to its high density of memristors. The capacity of the dense crossbar is easy to accommodate the size requirements of various neural network models, due to the high density of the memristors. The structure of the dense crossbar closely resembles a traditional memory design, as depicted in Figure~\ref{f:whole}(b). There are two significant features:

\vspace{0.1em}

\begin{enumerate}

\setlength\itemsep{0.15em}

\item Low-resolution DAC and ADC: Both the Digital-to-Analog Converter (DAC) and Analog-to-Digital Converter (ADC) employed in our system have low resolutions of 1 bit and 2 bits, respectively. The DAC functions by enabling or disabling the entire column of memristors, while the ADC incorporates a sense amplifier to recover the signal. Utilizing low-bit DAC and ADC resolutions significantly enhances energy efficiency and reduces the required area compared to higher-resolution alternatives~\cite{saberi2011analysis}.  

\item Individual column activation: At any given time, only one column of memristors is activated within the dense crossbar. This means that the current flowing through one memristor does not interfere with the current from other memristors in adjacent columns. This unique characteristic enables accurate data retrieval from large-scale crossbars without signal interference or degradation.

\end{enumerate}

\vspace{0.1em}
\noindent By incorporating the dense crossbar and computation crossbar within a single chip or package, we eliminate the inefficiency of off-chip communication. This design enables energy-efficient deployment of large-scale neural network models, particularly extremely large language models.
Furthermore, our dense crossbar provides comprehensive support for various configurations of memristors in any number of bits. In the specific example illustrated in Figure~\ref{f:whole}(b), we assume that each memristor can store two bits of information. This particular configuration represents a balance between the complexity of ADC and the additional area needed for implementing memristors. 

{Our architecture is compatible with a wide variety of memristors types. Memristors are implemented within the dense crossbar and arranged as a traditional memory bank. Considering the robustness of the dense crossbar in this particular organizational structure, the resolution and accuracy requirements for the memristors are relatively flexible. Therefore, our architecture can also support future advanced memristors with greater performance.
}

\begin{figure}[!t]
\centering
\includegraphics[width=3.5in]{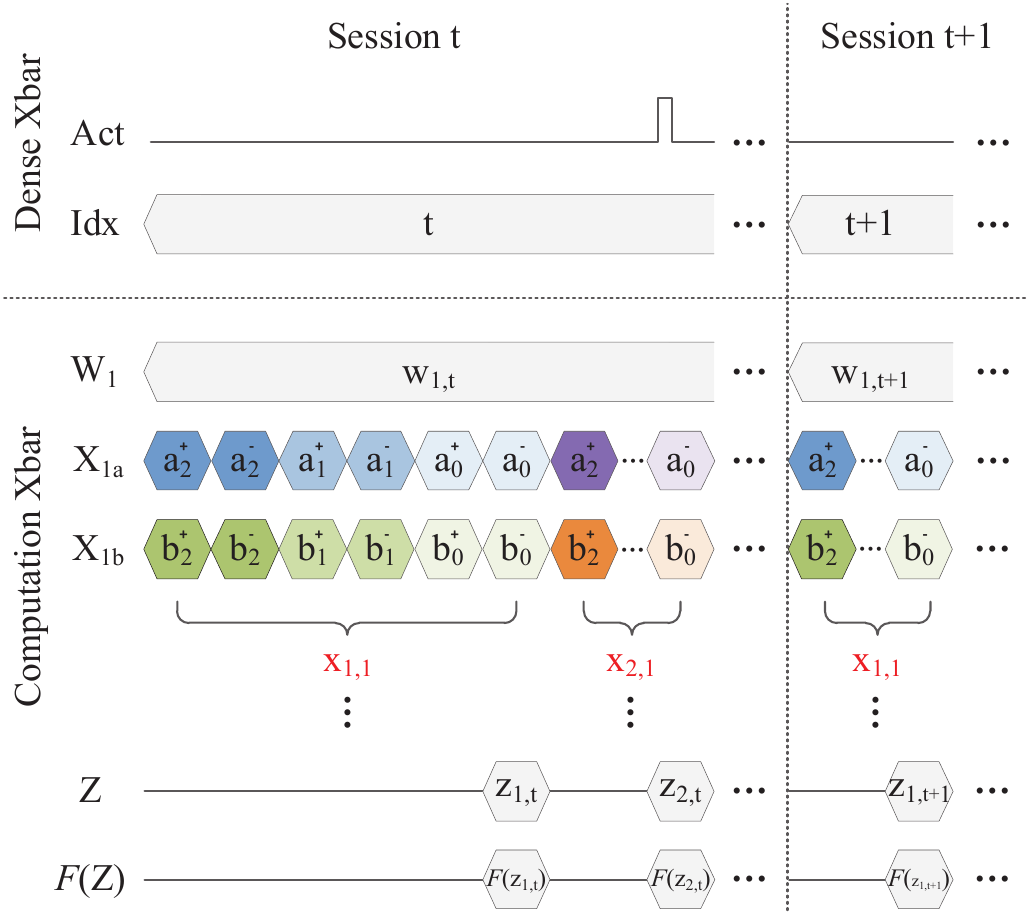}\\
\caption{The computation process of the sub-operation. Assuming W${_i}$ stores the weight, X${_i}$ stores the activation, and Z stores the computation results. The states of the four memristors are denoted as $a_i^+$, $a_i^-$, $b_i^+$, and $b_i^-$, where $i$ represents the index of the digital, and $a/b$ along with $+/-$ denote the location of the memristors.}
\label{f:wave}
\end{figure}


\subsection {Computation Process of Sub-Operation}

We illustrate the computation process of the sub-operation in Figure~\ref{f:wave}, depicting the sequential steps involved, assuming the linear part of the sub-operation is a weight-stationary multiplication. We use register W${_i}$ to store the weight, register X${_i}$ (including X${_{i\textrm{a}}}$ at row $a$ and X${_{i\textrm{b}}}$ at row $b$) to store the activation and register Z to store the computation result. The locations of these registers in the computation crossbar are shown in Figure~\ref{f:whole}(a). In this example, we adopt the balanced septenary (base-7) encoding system. Each activation $x_{i,j}$ (highlighted in red) is encoded into three digits. For each digit, we utilize four memristors. Two memristors in the positive column (storing $1$ in row $a$ and $2$ in row $b$) and two in the negative column (storing $-1$ in row $a$ and $-2$ in row $b$). The states of the four memristors are denoted as $a_i^+$, $a_i^-$, $b_i^+$, and $b_i^-$, where $i$ represents the index of the digital, and $a/b$ along with $+/-$ denote the location of the memristors. As three digits are required to encode each activation $x_{i,j}$, a total of $4 \times 3 = 12$ memristor states are required. During computation, we load the $12$ memristor states of activation $x_{i,j}$ into the respective registers X${_{i\textrm{a}}}$ and X${_{i\textrm{b}}}$ digit by digit.

As depicted by Equation~(\ref{e:unit2}), our proposed approach partitions the computation of sub-operations into multiple sessions. In session $t$, we load multiple weights from the dense crossbar to the computation crossbar. For instance, weight $w_{1,t}$ is loaded into {W}$_i$ and sequentially multiplied with activations $x_{1,1}$, $x_{2,1}$, and so on. As an example, the system performs the multiply–accumulate (MAC) operation by adding the product of $w_{1,t}$ and $x_{1,1}$ to other multiplication products. The column generates outputs $z_{1,t}$, $z_{2,t}$, and so forth. 
They are passed to the additional function block, denoted as $F$, and we obtain the final result $F(z_{1,t})$, $F(z_{2,t})$, and so on.
Once all possible activations have been traversed in session $t$, session $t+1$ begins. New weights are loaded from the dense crossbar to the computation crossbar and the same sequence of activations from session $t$ is repeated. This process continues for subsequent sessions. 

To expedite the computation time, we have employed a duplication technique for the computation columns associated with the same set of weights. This duplication approach significantly enhances the level of parallelism within each session. For instance, as illustrated in Figure~\ref{f:whole}(a), concurrently, we can calculate the multiplication between weights from W${_i}$ and activations from another activation register X${^{\prime}_i}$ in the second column of the computation crossbar. We load activations with even index $x_{2i,t}$ into X$_{1a}$ and X$_{1b}$, while activations with odd indices $x_{2i+1,t}$ are loaded into X$^\prime_{1a}$ and X$^\prime_{1b}$. This approach can reduce the processing time for each session by half. If the crossbar allows the implementation of $d_c$ columns, the processing time of each session can be further reduced to $1/d_c$ of the original value.

Our architecture is also capable of performing non-weight-stationary multiplication, which is essential for the multi-head attention block present in most language models.
As depicted in Figure~\ref{f:whole}, the multiplexers connecting the computation crossbar and dense crossbar play a crucial role in selecting the appropriate source for computation. They can choose between the weight values from the dense crossbar or another activation register {Y}$_i$. The computation process of non-weight stationary multiplication is similar to that of weight stationary multiplication.

\begin{figure}[!t]
\centering
\includegraphics[width=3.5in]{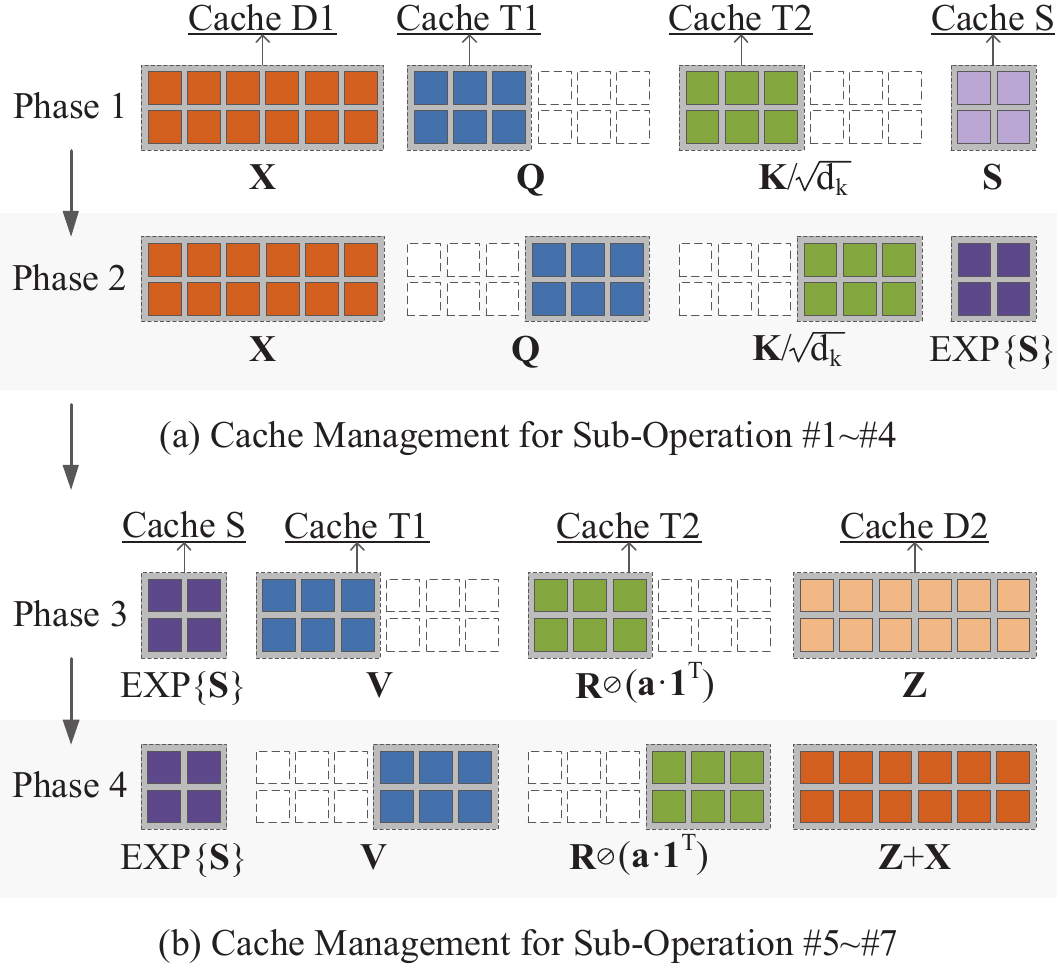}\\
\caption{{The cache management of the multi-head attention layer. Five caches are implemented, named Cache D1, D2, T1, T2, and S. Data stored within the caches are highlighted with colored blocks, while unstored data is represented by dashed blocks.}}
\label{f:cachemg}
\end{figure}

\subsection {{Cache Management System}}

{To temporarily store the intermediate results of the model inference, we need to implement five caches named Cache D1, D2, T1, T2, and S. When executing the multi-head attention (MHA) layers, T1 and D2 are used to store the input and output data, while caches T1 and T2 are employed to store temporary results. Cache S, on the other hand, is utilized to store the softmax result. When executing the feed-forward (FF) layers, only Cache D1 and Cache D2 are utilized to store the input and output data.}

{An example of the cache management flow for the MHA block is depicted in Figure~\ref{f:cachemg}. Similar approaches would be taken for cache management in the FF layers. Assuming each input sequence consists of two tokens, and each token has a dimension of $6$, a sequence can be represented as a $2\times6$ matrix. In this figure, data stored within the caches are highlighted with colored blocks, while unstored data is represented by dashed blocks. We divide the entire computation process into two parts. The first part in Figure~\ref{f:cachemg}(a) involves sub-operations 1$\sim$4, and the second part in Figure~\ref{f:cachemg}(b) involves sub-operations 5$\sim$7.}

{Cache D1 should have the capacity to store the entire matrix of the input sequence \textbf{X}. With the input data \textbf{X}, we can compute \textbf{Q}, \textbf{K}, and \textbf{V} column by column using our computing block. 
Cache T1 and T2 do not need to store the entire sequence, as subsequent operations can still be performed using partial columns from \textbf{Q} and \textbf{K}. In this example, Cache T1 and T2 are configured to store half of the columns in the sequence, which is why we need to undergo two phases to complete the computation of all columns. During each phase, the partially computed results of the softmax matrix \textbf{S} are accumulated in cache S. Until the final phase where the result is completed, we perform the additional exponential function on matrix \textbf{S} to obtain EXP(\textbf{S}).}
\renewcommand{\arraystretch}{1.2}

\begin{table}
\centering
\caption{{The cache size of our architectures. If the duplication factor exceeds $c_k$, then a cache capacity larger than the typical value is required.}}
\footnotesize
\begin{tabularx}{\linewidth}{c *{5}{|Y}  }
\toprule
Cache Size	& \$ D1	& \$ D2	& \$ T1	& \$ T2 & \$ S\\
\midrule
Typical	&$l_s\cdot d_k$	&$l_s\cdot d_k$	&$l_s\cdot c_k$ &$l_s\cdot c_k$	&$l_s\cdot l_s$	\\
Maximum	&$l_s\cdot d_k$	&$l_s\cdot d_k$	&$l_s\cdot d_k$  &$l_s\cdot d_k$ 	&$l_s\cdot l_s$	\\
\bottomrule
\end{tabularx}
\label{t:sizemg}
\end{table}

{Once the EXP(S) matrix is stored in cache S, we can clear the data stored in caches T1 and T2 and repurpose these two caches. Specifically, the computed matrix \textbf{V} is stored in cache T1, and the subsequent result \textbf{R} is stored in cache T2.
Similar to the previous part, we divide the computation process into two phases, and in each phase, only half of the columns in matrices \textbf{V} and \textbf{R} are stored in caches T1 and T2, respectively. During each phase, the partially computed results of the output matrix \textbf{Z} are accumulated in cache D2. Until the final phase where the result is completed, we perform the additional summation function on matrices \textbf{Z} and \textbf{X} to obtain matrix \textbf{Z+X}.}

{If we duplicate more computational crossbar units to achieve a speedup in computation, it is possible that cache T1 and T2 may not be sufficiently large to accommodate the intermediate computation results. In the extreme scenario, they should have the capacity to store the entire sequence of data, similar to how cache D1 and D2 do. To summarize, we have listed the typical and maximum sizes of each individual cache in Table~\ref{t:sizemg}. Here, the parameters $l_s$ and $d_k$ represent the number of tokens in the sequence and the dimension of each token, respectively. The parameter $c_k$ represents the number of columns that can be stored in cache T1 and T2. If the duplication factor exceeds $c_k$, then a cache capacity larger than the typical value is required.}

\section{Experiments}

Our assessment of LLM accuracy is based on PyTorch implementation using Hugging Face's package. RTN Fluctuation functions are applied to activations and weights during computation to simulate device noises. Quantization functions are applied to simulate real device conditions. To ensure fair comparisons, we adjust the resistance range of memristors for each crossbar architecture, guaranteeing equal accuracy levels across architectures. Accuracy evaluation is conducted on language tasks from the GLUE dataset~\cite{wang2018glue}. We fine-tuned the pre-trained models for $5$ epochs on tasks SST-2, QQP, MNLI, and QNLI. For the remaining tasks, which are relatively small, we fine-tuned them for $10$ epochs. The batch sizes for BERT$_\textrm{Base}$ and  BERT$_\textrm{Large}$ are $32$ and $16$. The experiments are performed on 2080TI GPU cards, with each experiment completed within one day. 

We use simulation tools~\cite{lee2019system,luo2018fpga} to evaluate area overhead, energy consumption, and latency. {The simulator includes noise models and non-ideality models for memristors. The noise parameters utilized in the simulation are derived from measured data obtained from real fabricated memristor devices~\cite{hu2016dot}.} These tools are built with synthetic data from EDA tools~\cite{lavagno2018eda} and calibrated using experimental data from real chips called Novena~\cite{9336142}. {Table~\ref{t:spec} provides a list of the hardware parameters employed in the simulation. This includes our settings for DAC, ADC, memristor, and register on a per-crossbar basis, as well as the configurations of caches at the architectural level.}

\renewcommand{\arraystretch}{1.2}

\begin{table}
\centering
\caption{{Hardware parameters used in the simulations. The cache parameters apply across the entire architecture level.}}
\footnotesize
\label{t:spec}
\begin{tabularx}{\linewidth}{l*{4}{|Y} Y}
\toprule
 &\multicolumn{2}{c}{Per Computation Crossbar}        &  \multicolumn{2}{|c}{Per Dense Crossbar}   \\ 
\cmidrule{2-5}
  &Type     & Qty.     &Type     & Qty.   \\ 
\midrule
DAC       & 8-bit & $\times$128  & 1-bit & $\times$64$k$  \\
ADC       & 8-bit & $\times$1  & 2-bit & $\times$1$k$  \\
Memristor      & Resistor & 128 $\times$128 & Regular & 1$k$$\times$64$k$ \\
Register       &6-bit & $\times$128$\times$64  & 8-bit & $\times$256   \\
\midrule
Cache$^{1}$  &\multicolumn{4}{l}{\  D1:256kB \quad D2:256kB \quad  T1:16kB \quad  T2:16kB  \quad  S:64kB}  \\
\bottomrule
\multicolumn{5}{l}{$^1$\ Assuming sequence length of 256 tokens with 1024 dimensions}
\end{tabularx}
\end{table}

\renewcommand{\arraystretch}{1.2}

\begin{table}
\centering
\caption{Comparison among base values, assuming the required data precision for activations is INT8 (i.e., 8-bit integer)}
\footnotesize
\label{t:base}
\begin{tabularx}{\linewidth}{l  *{7}{|Y} Y }
\toprule
Base Value &3 &7 &15 &31 &63 &127 &255             \\ 
\midrule
Scale Factor S       &1  &2  &3  &4  &5  &6  &7     \\
Digits/Cycles               &6  &3  &3  &2  &2  &2  &1     \\
Scale-Cycle Product  &6  &6  &9  &8  &10 &12 &7     \\ 
\bottomrule
\end{tabularx}
\end{table}

We conducted tests on various Language Models (LLMs) including BERT, Phi-1.5~\cite{textbooks2}, GPT-2, T5, LLaMa, and GPT-3 to evaluate the performance of the crossbars. In addition to LLMs, we also evaluated the CV models (ResNet). We assume these models operate with 8-bit activations and 8-bit weights. The sizes of the multi-bit, single-bit, and computation crossbars are assumed to be $128\times 128$. The size of the dense crossbar is $1k\times 64k$, matching that of 
a DRAM bank~\cite{dram}. Following the setting in ISAAC, each ADC is shared among 128 columns as its switching speed is 128 $\times$ faster than the memristors~\cite{shafiee2016isaac}.

We follow the evaluation methodology employed in three highly cited works: PRIME~\cite{chi2016prime}, ISAAC~\cite{shafiee2016isaac}, and PipeLayer~\cite{song2017pipelayer}. To make a fair comparison of area overhead and energy consumption, we utilize the circuit components from ISAAC~\cite{shafiee2016isaac} as a basis for all the architectures. {In our experiments, we override the device parameter file in the simulator to update the respective parameters.} The crossbars consist of DACs, ADCs, registers, memristors, and peripheral circuits such as sample-and-hold, shift-and-add, encoder, and the operation unit $F$. 

For the DACs, we rely on the analysis presented in~\cite{saberi2011analysis} to determine their area and energy consumption. The ADC model is based on~\cite{kull20133}. Following the approach used in ISAAC, we scale the area and energy of single-bit DACs and 2-bit ADCs using the analytical models from~\cite{saberi2011analysis}. The sample-and-hold circuit data is sourced from~\cite{o200410}. The area and energy consumption of the shift-and-add circuit is determined based on the analysis conducted in DaDianNao~\cite{chen2014dadiannao}. In the experiment, we choose to utilize a digital circuit to implement function $F$. Metrics on the unit $F$ and the encoder are retrieved from the synthesis report of EDA tools~\cite{lavagno2018eda}. We employ the CACTI 6.5~\cite{muralimanohar2007optimizing} tool to model the energy and area of registers. The energy and area model for memristors is derived from~\cite{hu2016dot}. All these components are assumed to be fabricated under the 32 nm node, the same as ISAAC.

{The memristors we employed in our experiments is a TaOx-based device, demonstrating approximately 1\% error rate when configured with $32$ (5-bit) conductance levels~\cite{hu2016dot}. This is equivalent to nearly 100\% accuracy when programmed with just $4$ conductance levels (2-bit) within our dense crossbar. We chose this device based on the specifications outlined in ISACC~\cite{shafiee2016isaac}, in order to establish a fair comparison between our architecture and theirs. It offers a precision level sufficient to accommodate our architectures and most other architectures, such as PRIME~\cite{chi2016prime} and PipeLayer~\cite{song2017pipelayer}, ensuring reasonable model accuracy.}

\begin{figure}[!t]
\centering
\includegraphics[width=3.5in]{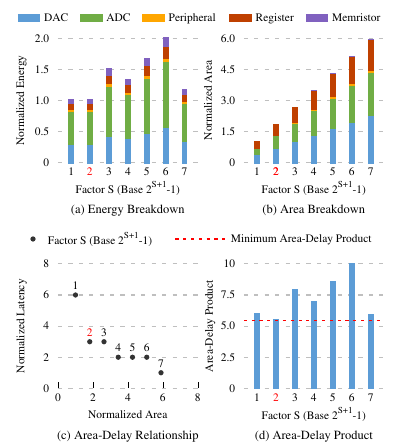}\\
\caption{Comparison among various scaling factors $S$ (base value $=2^{S+1}-1$), assuming the required data precision for activations is INT8 (i.e., 8-bit integer): (a) Energy breakdown of the computation crossbar; (b) Area breakdown of the computation crossbar; (c) Area-latency distribution; (d) Area-delay product.}
\label{f:bases}
\end{figure}

\renewcommand{\arraystretch}{1.7}

\begin{table*}[!t]
\caption{The comparison of accuracy between our architecture and the baseline on the GLUE tasks. Simulation results show that the environmental noise contributes less than 5\% to the signal level, which is typical for real devices~\cite{feinberg2018making}. Weights and activations are quantized into 8 bits.}
\resizebox{1.00\linewidth}{!}{
\begin{tabular}{l | c |c| c  c| c c|  c c| c  c| c| c|  c  }
\toprule
&CoLA & SST-2 &\multicolumn{2}{c|}{MRPC} &\multicolumn{2}{c|}{STS-B} &\multicolumn{2}{c|}{QQP}&\multicolumn{2}{c|}{MNLI} & QNLI & RTE & WNLI \Tstrut \\										& M.C.	& Acc.	& F1	& Acc.	& Pea.	& S.C.	& Acc.	& F1	& Acc.	& MM.	& Acc.	& Acc,	& Acc.	\Bstrut\\
\hline
BERT$_\textrm{Base}$ - {Baseline} &58.03	&93.00	&88.39	&83.58	&88.96	&88.48	&91.00	&87.80	&83.68	&83.54	&90.66	&65.34	&30.99	\Tstrut  \\
BERT$_\textrm{Base}$ - \textbf{Ours } &59.07	&93.12	&89.20	&85.05	&88.92	&88.45	&90.91	&87.80	&83.77	&83.53	&90.57	&62.45	&30.99	\Bstrut \\
\hline	
BERT$_{\textrm{Large}}$ - {Baseline} &62.63	&92.43	&91.54	&87.99	&89.72	&89.53	&91.22	&88.17	&86.01	&86.11	&92.37	&68.23	&57.75	\Tstrut  \\
BERT$_{\textrm{Large}}$ - \textbf{Ours }&62.43	&92.78	&91.77	&88.48	&89.68	&89.47	&91.13	&88.16	&86.23	&86.15	&92.22	&68.23	&57.75	\Bstrut \\
\bottomrule
\end{tabular}
}
\label{t:accuracy}
\end{table*}

\begin{table}[!t]
\caption{{The running accuracy of Phi-1.5~\cite{textbooks2} and GPT-2 models on our architecture. The label (3-c) refers to 3-cycle computation for each column (same as other experiments), and (4-c) refers to 4-cycle.}}
\resizebox{1\linewidth}{!}{
\begin{tabularx}{\linewidth}{l |*{3}{Y}| *{3}{Y} Y }
\toprule
 &\multicolumn{3}{c|}{Phi-1.5}   &\multicolumn{3}{c}{GPT-2} \\
\cmidrule{2-7}
	& Base-	&\textbf{Ours}	& \textbf{Ours} 
 & Base- &\textbf{Ours}	&\textbf{Ours}	 \\
Task list &line & (4-c) & (3-c) &line & (4-c) & (3-c)\\
\midrule
WinoGrande	&0.729	&0.729	&0.717	&0.516	&0.519	&0.51	\\
ARC\_Easy	&0.762	&0.758	&0.756	&0.438	&0.439	&0.406	\\
ARC\_Challenge	&0.445	&0.45	&0.451	&0.19	&0.183	&0.195	\\
PIQA	&0.766	&0.762	&0.755	&0.629	&0.637	&0.599	\\
Hellaswag	&0.48	&0.473	&0.473	&0.289	&0.29	&0.292	\\
MMLU	&0.418	&0.398	&0.394	&0.229	&0.229	&0.229	\\
OpenbookQA	&0.386	&0.39	&0.392	&0.164	&0.152	&0.15	\\
\bottomrule
\end{tabularx}
}
\label{t:accuracy_real}
\end{table}

In our evaluation, we compare our work with these three state-of-the-art RRAM solutions: PRIME, ISAAC, PipeLayer, as well as an area-efficient SRAM-based crossbar called Vesti~\cite{yin2019vesti}. It is important to note that these architectures do not support non-weight stationary multiplications. Therefore, our comparison focuses solely on the weight stationary computation aspect. {The calculated metrics for PRIME, ISAAC, and PipeLayer only includes modules for WS operations. Our architecture encompasses both the computation and dense crossbars so that we can execute the same set of operations, .}

Additionally, we compare our work with Google's TPUv4 accelerator and Nvidia's A100 GPU, which represent state-of-the-art high-performance solutions for machine learning applications. During the accuracy evaluation of LLM on language tasks, TPU/GPU operates in full precision mode, while our architecture utilizes quantized weights and activations.

\subsection{Searching for the Optimal Encoding Base}

Our system offers compatibility with various encoding bases, and we can utilize grid search to determine the optimal encoding base that can achieve the highest efficiency under the required data precision for activations. In Table~\ref{t:base}, we present a comparison of different encoding bases, assuming the required data precision for activations is INT8 (i.e., 8-bit integer). The table clearly demonstrates that utilizing a larger base value enables the encoding of the same data into fewer digits, thereby reducing the number of cycles required for computation. However, the number of resistors required to encode the activation (equalling $2S$) is increased with a larger base. To ensure a fair comparison across different base values, we introduce a metric called the scale-cycle product, which indicates the minimum latency achieved with one unit of memristors. For instance, in the base-7 encoding system with a scale value $S=2$, we can use $2\times$ resistors to achieve a 3-unit latency. Therefore, the scale-cycle product is $6\times$, indicating that the latency would be 6 units if we only had $1\times$ resistor. As we aim for a lower scale-cycle product, the base-3 $(S=1)$ and base-7 $(S=2)$ emerge as the optimal choices.

In Figure~\ref{f:bases}(a) and Figure~\ref{f:bases}(b), we present the energy and area breakdown of the computation crossbar. The area of the dense crossbar remains the same across different encoding schemes as we store the entire network parameters. Therefore, our main focus is comparing the encoding schemes on the computation crossbar. The analysis reveals that a significant portion of the energy consumption and area overhead on the computation crossbar is attributed to the DACs and the ADCs.
Among all the base values, factors $S = 1$ (base-3) and $S = 2$ (base-7) exhibit the lowest energy consumption primarily due to their low scale-cycle product. On the other hand, the area overhead experiences an almost linear increase with the scaling factor $S$. This is because the number of memristors is proportional to the factor $S$. 
Furthermore, Figure~\ref{f:bases}(c) presents the area-latency distributions, while Figure~\ref{f:bases}(d) showcases the area-delay product (ADP). These figures demonstrate that a larger scaling factor $S$ results in lower latency; however, it does not correspond to a lower ADP. Notably, when comparing factor $S = 2$ with factor $S = 1$, about 9\% reduction in ADP is observed. The variations primarily stem from the differences in the registers employed. 

\begin{figure*}[!t]
\centering
\includegraphics[width=7.15in]{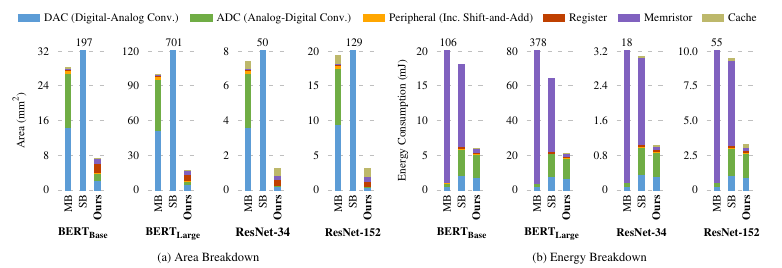}\\
\caption{The comparison among the multi-bit (MB) crossbar, the single-bit (SB) crossbar, and our work in (a) area overhead, and (b) energy consumption. Some results exceed the upper bound of the y-scale. In that case, we mark the exact value of the metric on top of the bar.}
\label{f:compare}
\end{figure*}

\subsection{Accuracy/Scores on Language Tasks}

Compared to digital circuits, analog circuits are more vulnerable to environmental effects such as noise~\cite{raghavan2013rtn}. Additionally, when using quantized weights and activations, memristors lose precision compared to the full precision version, potentially affecting the accuracy of models. To assess the robustness of our architecture, we simulated the environment~\cite{lee2019system,luo2018fpga} and tested our architecture on the GLUE language tasks~\cite{wang2018glue} using BERT models. 
Simulation results show that the environmental noise contributes less than 5\% to the signal level, which is typical for real devices~\cite{feinberg2018making}. The results are summarized in Table~\ref{t:accuracy}, demonstrating that our architecture performs almost as well as {baseline} from GPU on both BERT$_\textrm{Base}$ and BERT$_\textrm{Large}$ applications across all tasks in the GLUE benchmark. In some cases, we even observed a slight increase in accuracy. This can be attributed to the errors in the results. {We also testing real world datasets on Phi-1.5~\cite{textbooks2} and GPT-2. The experiments results in Table~\ref{t:accuracy_real} demonstrate that our system achieved similar accuracy and scores to the baseline.}

Based on the two tables, it can be inferred that the amount of noise has a negligible impact on the model's accuracy. {The robustness of our dense crossbar plays a role in stabilizing model accuracy.
To assess the robustness of our architecture, we conducted tests in even more challenging noise environments. Experiments show that with a noise amplitude 5X stronger, the output from the dense crossbar and the accuracy of the model remain unaffected. Moreover, if we choose to program each memristor with 1-bit information using two conductance levels, our system can withstand even stronger noise environments, up to 18X the noise amplitude.} Additionally, we believe that the softmax function within LLMs also plays a crucial role in enhancing noise tolerance as it stops the accumulation of calculation errors. This advantage sets LLMs apart from applications that demand high precision, making them less susceptible to the negative effects of environmental noises. 

\subsection{Reduction of Area Overhead}

{To reduce the number of ADCs/DACs and consequently decrease the crossbar area, an effective approach is to increase the number of rows and columns in a computation crossbar, allowing a single ADC or DAC to be shared by more memristors. However, this is challenging due to the accumulation of noise from the non-ideal behavior of memristors~\cite{bhattacharjee2023examining}\cite{9218688}\cite{8326998}.
On the other hand, memristors assembled like a traditional memory bank can have a much larger size as they are more resilient to noise. As each memristor in the crossbar works independently, errors do not accumulate over the columns~\cite{itoh2001high}\cite{xu2015overcoming}. With ADCs and DACs shared by more memory cells, this type of architecture has higher area efficiency than traditional computation crossbars. 
Our architecture improves area efficiency by combining these two types of crossbars. As illustrated in Figure~\ref{f:overview1}, the computation crossbar continues to use the classical design for analog computing with a small crossbar dimension ($128\times 128$). Model parameters are stored in the dense crossbar, which employs the traditional memory bank design with a significantly larger crossbar dimension ($1k\times 64k$). We enable the reconfiguration of the computation crossbar and transfer the weights from the dense crossbar.}

As illustrated in Figure~\ref{f:compare}(a), our architecture significantly reduces the area requirement,  On average, our architecture achieves approximately {$6\times$} and {$39\times$} savings in area overhead compared to the multi-bit architecture and single-bit architecture, respectively. We tune the configuration (duplicated columns in the computation crossbar) of our architecture to guarantee that the end-to-end delay of our architecture is equivalent to these two architectures. In either the multi-bit or the single-bit version, all network parameters are stored within conventional crossbars. As shown in the figure, the effective density of conventional memristor crossbars is relatively low, with approximately {95\%} of the area occupied by DACs and ADCs. In contrast, our architecture stores all the parameters in the dense crossbar. Due to the independent nature of data read by each memristor, which does not impact other memristors, the DACs and ADCs in our dense crossbar can be shared among a larger number of memristors compared to the conventional crossbar~\cite{dram}. Furthermore, the 1-bit DACs and 2-bit ADCs occupy a considerably smaller portion of the overall area. These features allow us to deploy even larger LLM on a single chip with a lower area overhead, thus eliminating the time and energy inefficiencies associated with off-chip communication. 
\begin{figure*}[!t]
\centering
\includegraphics[width=7.15in]{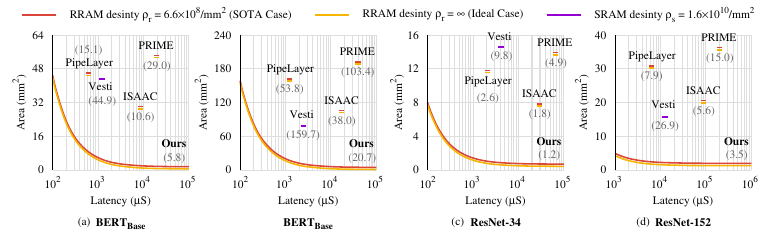}\\
\caption{{(a-d) Comparison with the state-of-the-art {RRAM} architectures: (\textbf{P}) PRIME~\cite{chi2016prime}, (\textbf{I}) ISAAC~\cite{shafiee2016isaac}, (\textbf{L}) PipeLayer~\cite{song2017pipelayer}, and an area-efficient {SRAM} architecture (\textbf{V}) Vesti~\cite{yin2019vesti}. We label the energy consumption (\textit{mJ}) in the brackets. (\textit{e.g.}, I(10.6) indicates an ISAAC crossbar consuming 10.6 mJ energy.) All the architectures are adapted to 8-bit activations/weights under the 32 nm technology node, the same as ISAAC~\cite{shafiee2016isaac}.}}
\label{f:sota}
\end{figure*}

\subsection{Reduction of Energy Consumption}

{In traditional memristor-based crossbar designs, a significant amount of energy is consumed by memristors~\cite{murali2020heterogeneous}\cite{wang2023optimizing}. One major drawback of memristors is their inherent issues with non-ideality and noise, particularly the random telegraph noise (RTN) ~\cite{tseng2010electron}\cite{terai2009effect}, which arises as unresolved defects during programming. The amplitude of the fluctuation is roughly proportional to the resistance levels~\cite{ielmini2010resistance}\cite{lee2011noise}. To alleviate the negative impact of the fluctuations on the model, developers need to decrease the resistance to reduce the fluctuations~\cite{zanotti2021low}\cite{chai2018impact}. However, smaller resistance values result in larger currents, leading to higher energy consumption~\cite{wang2023optimizing}\cite{gao2013digital}.
We re-architected the crossbars to enhance their robustness against fluctuations in memristors. First, as our algorithms only require them to store the same data value during computation, the memristors in the computation crossbar are replaced with regular resistors. Without the programming process, the resistor exhibits significantly smaller fluctuations~\cite{kayanalysis}\cite{ielmini2010resistance}. Secondly, the memristors in the dense crossbar function like traditional memory bank. Therefore, memristor errors do not accumulate over the column and can be recovered by the output circuit as long as the fluctuations do not exceed the threshold~\cite{nam2014quantitative}\cite{seyedzadeh2017mitigating}.}

As demonstrated in Figure~\ref{f:compare}(b), our architecture achieves significant energy savings. On average, our architecture achieves approximately {$18\times$} and {$3\times$} reductions in energy consumption compared to the multi-bit architecture and single-bit architecture, respectively. The figure reveals that the multi-bit architecture consumes more energy due to the need for memristors to operate in the low-resistance mode to counter noise, particularly the RTN (random telegraph noise) effects~\cite{raghavan2013rtn}. In contrast, the single-bit architecture enhances memristor robustness by storing only one-bit information, allowing for higher resistance levels and lower energy consumption~\cite{zhu2019configurable}.
In our computation crossbar, we utilize regular resistors as replacements for memristors, leveraging their greater physical robustness against RTN noise~\cite{ielmini2010resistance}. Additionally, our dense crossbar incorporates two key mechanisms to enhance robustness. First, the data read by each memristor of every column is independent and is does not affected by other memristors in the columns~\cite{dram}. Second, we employ 1-bit DACs and 2-bit ADCs in the input/output circuits, which are more robust than multi-bit DACs and ADCs~\cite{dram}. The incorporation of these robustness mechanisms allows the resistors and memristors in our crossbars to operate in the low-resistance mode, leading to reduced energy consumption. 

\renewcommand{\arraystretch}{1.2}

\begin{table}
\centering
\caption{Comparison between TPUv4, A100, and our architecture on BERT$_{\textrm{Large}}$, adapting to INT8 operations under 32 nm technology node.}
\footnotesize
\begin{tabularx}{\linewidth}{ *{3}{Y|}  *{3}{|Y}   }
\toprule
\multicolumn{3}{c||}{ADP (mm$^2\cdot$ s)} &  \multicolumn{3}{c}{Energy (mJ)}   \\ 
\midrule
TPU &  GPU & Ours  &TPU &GPU &Ours  \\ 
2.24		&2.04		&0.03		&71.71		&65.84		&20.44	\\
\bottomrule
\end{tabularx}
\label{t:compare_table}
\end{table}

\subsection{Comparison with the State-of-the-Art}

In Figure~\ref{f:sota} (a-d), we present a comparison of our architecture with three popular architectures for RRAM memristor crossbars: \textbf{P}RIME~\cite{chi2016prime}, \textbf{I}SAAC~\cite{shafiee2016isaac}, and Pipe\textbf{L}ayer~\cite{song2017pipelayer}. These architectures represent three distinct strategies aimed at reducing the area of the memristor crossbar. PRIME utilizes a sense amplifier and a dynamic reference voltage source to convert analog data into digital values. ISAAC employs 1-bit DACs as inputs and accumulates results over multiple time steps. PipeLayer eliminates the need for ADCs by transforming analog signals into spikes using capacitors.  
The comparison is based on factors such as area, latency, and energy consumption (values provided in brackets). For the purpose of comparison, we adapted each architecture's design to accommodate 8-bit weights/activations. In contrast, our architecture is represented by a curve in the graph since we offer different solutions by adjusting the number of duplicated columns in the computation crossbars. This flexibility allows us to tailor the system to specific requirements and achieve optimal results.

Our system showcases substantial advancements in various aspects, including reductions in area overhead, energy consumption, and latency. The enhanced computation parallelism achieved by duplicating more columns in the computation crossbar for the same set of weights contributes to its low latency. 
{To clearly demonstrate the effectiveness of our architecture and eliminate the impact of memristor improvements on area efficiency, we analyze two cases for the memristor-based crossbars in Figure~\ref{f:sota}. In the first case, all the architectures are compared based on memristors with state-of-the-art density, shown as a purple line. In the second case, we assume the density of the memristors is infinitely high, shown as a yellow line. As we can see, even when we eliminate the impact of memristors, our architecture still exhibits better performance than the traditional one in terms of area.}

In Figure~\ref{f:sota} (a-d), we also compare our work with Vesti~\cite{yin2019vesti}, an SRAM-based system that improves area efficiency by reusing its SRAM crossbar. Our system exhibits substantially lower energy consumption compared to Vesti, primarily due to reduced off-chip communication requirements.
Furthermore, in Table~\ref{t:compare_table}, we compare our work with state-of-the-art TPU accelerators and GPUs that utilize HBM (High Bandwidth Memory) to address the memory wall problem~\cite{wang2020shuhai}. Leveraging the analog computing features, our system outperforms these alternatives, achieving a minimum of $68\times$ improvement in ADP (Area-Delay Product) and 69\% energy savings on BERT$_{\textrm{Large}}$.

\renewcommand{\arraystretch}{1.2}

\begin{table}
\centering
\caption{{Latency lower bounds of our architecture, utilizing the same experimental setup as shown in Figure~\ref{f:sota}.}}
\begin{tabularx}{\linewidth}{c *{4}{|Y} Y }
\toprule
Bound ($\mu$s) &  BERT$_\textrm{Base}$ & BERT$_\textrm{Large}$  &ResNet-34 &ResNet-152  \\ 
\midrule
$T^a_{\textrm{LB}}$	&16.1	&43.0	&2.7	&19.4	\\
$T^w_{\textrm{LB}}$ &17.3	&61.5	&4.3	&11.2	\\
\bottomrule
\end{tabularx}
\label{t:bandmg}
\end{table}

{Even without the impact of NW and non-linear operations, our architecture still exhibits advantages in terms of area and energy.
To enable these operations, one possible solution for traditional architectures is to incorporate specialized calculation units alongside the memristor crossbar. This additional module further widens the area gap between our architecture and the traditional ones.
For NW operations, our architecture continues to exhibit lower energy consumption per operation compared to traditional digital circuits. For example, approximately, the energy efficiency of TPUv4~\cite{jouppi2023tpu} at 7nm technology is 170 watts per 275 trillion INT8 operations, which equals approximately 0.62 pJ/OP. In contrast, our architectural experiments reveal an average energy consumption of 20.7 mJ for 77.3 operations, indicating an energy efficiency of 0.27 pJ/OP at the 32 nm technology node.}

\subsection{{Inference Latency Lower Bound}}

{The inference latency is constrained either by the minimum transmission time of activations from the cache, denoted as $T^a_{\textrm{LB}}$, or by the minimum transmission time of weights from the dense crossbar, denoted as $T^w_{\textrm{LB}}$. With these two parameters, we can estimate the lower bound of the inference latency using Equation~(\ref{e:lowerb}).}
\begin{equation}
{T_{\textrm{LB}}= \textrm{max}\{T^a_{\textrm{LB}}, T^w_{\textrm{LB}} \} = \textrm{max}\{\frac{ \alpha_a \cdot S_a  \cdot b_a }{ B_a}, \frac{N_w\cdot b_w }{ B_w}\}}
\label{e:lowerb}
\end{equation}

{To calculate $T^a_{\textrm{LB}}$, we need to estimate the total amount of transmitted data from the cache. This can be achieved by multiplying three variables: $\alpha_a$, representing the activation read times, $S_a$, denoting the number of elements in the activation data, and $b_a$, which indicates the bitwidth of each activation element. Next, we divide the data size by the on-chip communication bandwidth from the cache, denoted as $B_a$. In this experiment, we assume $B_a$ to be $1000$ GBps~\cite{Intel}, which is a typical bandwidth of L3 cache in modern CPUs, whose capacity is large enough to hold our intermediate results.}

{We can employ similar approaches to calculate $T^w_{\textrm{LB}}$. The total size for weight transmission is represented as $N_w\cdot b_w$, where $N_w$ denotes the number of parameters in the model, and $b_w$ stands for the bitwidth of the parameter. This total size is subsequently divided by $B_w$, which denotes the bandwidth between the computation crossbar and the dense crossbar. We refer to the bandwidth value $B_w$ from the HBM3 standard, assuming it to be $819$ Gbps per stack~\cite{Hynix}.}
{In Table~\ref{t:bandmg}, we provide the lower bounds of latency associated with the experiment in Figure~\ref{f:sota}. As these lower bounds extend beyond the X-axis range, they are not explicitly represented in the figure.}

\subsection{Scalability Analysis on Larger Models}

In recent times, there has been a rapid surge in the size of large language models (LLMs), leading to a notable escalation in the area overhead of traditional memristor crossbars. This is due to the need for a larger number of memristors to store all the model parameters. The physical limitations make it challenging to deploy the model on a single-chip system thus avoiding the inefficiency caused by off-chip communications. Although the utilization of 3D stacking techniques~\cite{oota20193d} aids in alleviating this concern, the overall area overhead within one chip or package is still limited. Hence, the growing number of parameters in state-of-the-art LLMs presents a challenge for adopting LLMs in memristor crossbars. The objective of this experiment is to thoroughly analyze and evaluate the scalability of our memristor crossbar architecture on both current and upcoming LLMs, and to determine whether we can successfully deploy them on a single chip or package.

\begin{figure}[!t]
\centering
\includegraphics[width=3.5in]{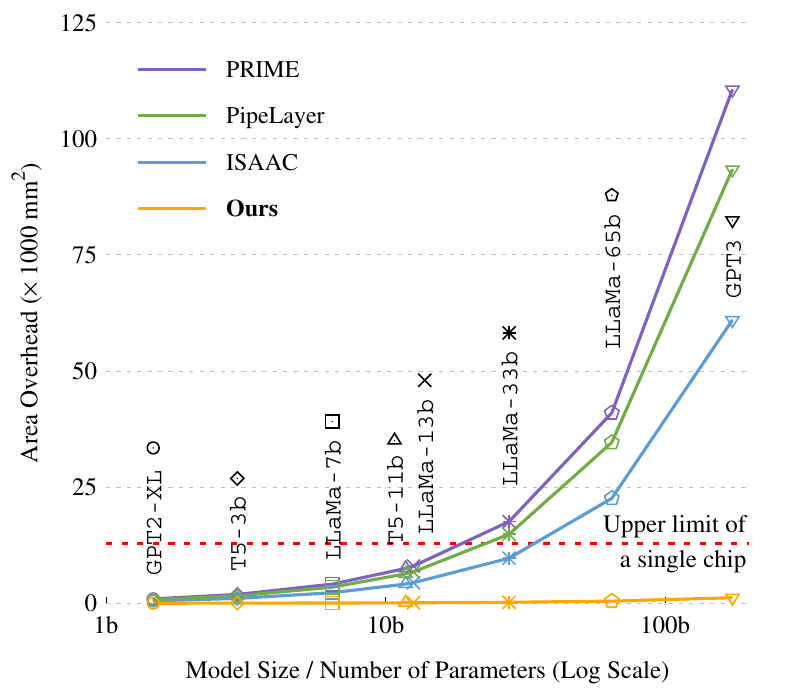}\\
\caption{The area overhead of state-of-the-art architectures and our architecture under large-scale LLMs. They are compared on the same 32 nm technology node with ISAAC~\cite{shafiee2016isaac}. The dashed line represents the upper limit of a single chip (assuming $100 mm^2\times 128$ layers)}
\label{f:sota2}
\end{figure}

In Figure~\ref{f:sota2}, we compare the area overhead of our architecture with state-of-the-art architectures under various LLMs, including GPT-2, T5, LLaMa, and GPT-3. They are compared on the same 32 nm technology node with ISAAC~\cite{shafiee2016isaac}.
The figure clearly illustrates that as the model size expands, our architecture showcases a significantly lower increase rate in area overhead compared to previous architectures, which experience rapid growth in area overhead. As an example, when considering the deployment of GPT-3, our architecture demonstrates an area overhead that is merely 1/51th of the area occupied by the previous architecture. Based on the observed trend, it is anticipated that future LLMs with even more parameters can be deployed in our architecture within one chip or package under a reasonable area overhead. 

\section{Conclusion}

This paper introduces a novel architecture of memristor crossbar that enables the deployment of state-of-the-art LLM on a single chip or package, effectively bypassing the energy and time inefficiencies associated with off-chip communication.
It addresses three significant challenges encountered when depolying LLMs on memristor crossbars, namely the large model size, the non-weight stationary multiplication, and the complex non-linear operations, which have traditionally posed significant obstacles for memristor crossbars.
The introduced architecture demonstrates substantial improvements in both area and energy efficiency. After testing BERT$_{\textrm{Large}}$, we found that our architecture incurred negligible accuracy loss. In comparison to traditional memristor crossbars, our design offers remarkable improvements, with up to {$39\times$} reduction in area overhead and {$18\times$} reduction in energy consumption. When compared to modern TPU/GPU systems, our architecture achieves a minimum of $68\times$ reduction in the area-delay product and significantly lowers energy consumption by 69\%. Furthermore, we observe a $51\times$ improvement in area overhead for GPT-3.

{\footnotesize\bibliography{IMC}
\bibliographystyle{IEEEtran}
}

\begin{IEEEbiography}[{\includegraphics[width=1in]{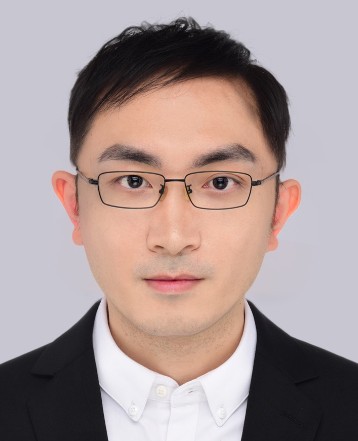}}]{Zhehui Wang}
received B.S. degree in Electrical Engineering from Fudan University, China, in 2010, and Ph.D. degree in Electronic and Computer Engineering from Hong Kong University of Science and Technology, Hong Kong, in 2016. He is currently a Research Scientist with the Institute of High Performance Computing, Agency for Science, Technology and Research,
Singapore. He authored and co-authored more than 60 research papers in peer-reviewed journals, conferences, and books. His research interests include efficient AI deployment, AI on emerging technologies, hardware-software co-design, and high-performance computing.
\end{IEEEbiography}


\begin{IEEEbiography}[{\includegraphics[width=1in]{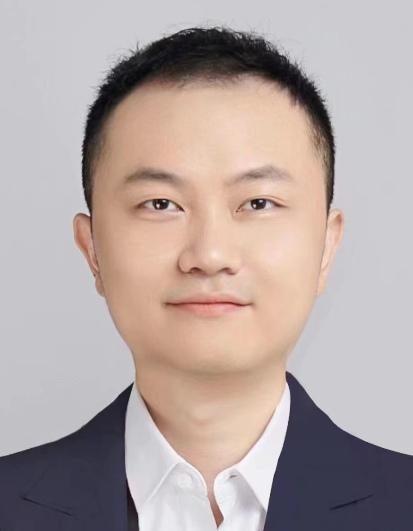}}]{Tao Luo}
received his bachelor’s degree from the Harbin Institute of Technology, Harbin, China, in 2010, his master’s degree from the University of Electronic Science and Technology of China, Chengdu, China, in 2013, and his Ph.D. degree from the School of Computer Science and Engineering, Nanyang Technological University, Singapore, in 2018. He is currently a senior research scientist with the Institute of High Performance Computing (IHPC), Agency for Science, Technology and Research, Singapore (A*STAR), Singapore. He has authored over 50 scientific publications in premier peer-reviewed international conferences and journals. His current research interests include high-performance computing,  machine learning, hardware–software co-exploration, quantum computing, efficient AI and its application.
\end{IEEEbiography}


\begin{IEEEbiography}[{\includegraphics[width=1in]{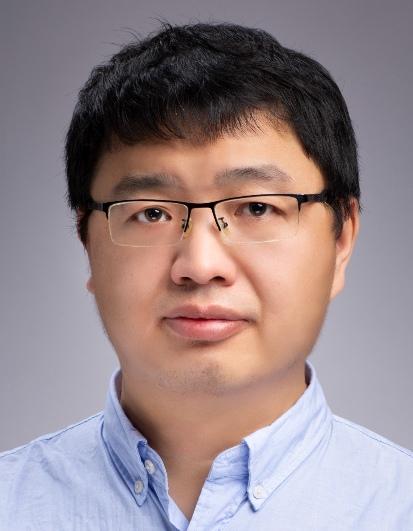}}]{Cheng Liu}
is as an Associate Professor at the State Key Lab of Processors (SKLP), Institute of Computing Technology (ICT), Chinese Academy of Sciences (CAS). He received BEng and MEng degrees from Harbin Institute of Technology, Harbin, China, in 2007 and 2009 respectively, and the PhD degree from the University of Hong Kong, in 2016. His research interests include fault-tolerant computing, domain specific architectures, computing in memory, and AI4EDA. He has authored over 70 scientific publications in premier international conferences and journals. He won the Best Paper Award at the Great Lakes Symposium on VLSI in 2021 and IEEE Transactions on Computers in 2021. He is a recipient of Huawei OlympusMons Awards in 2024.
\end{IEEEbiography}

\newpage
\begin{IEEEbiography}[{\includegraphics[width=1in]{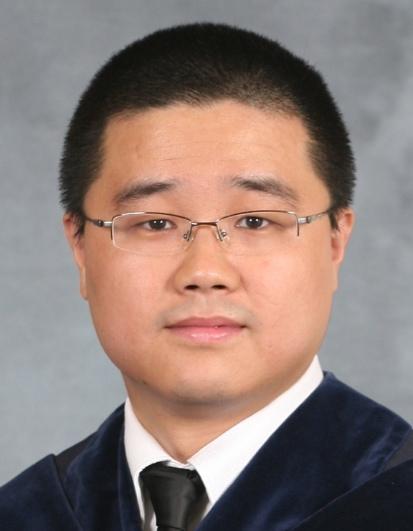}}]{Weichen Liu}
received his BEng and MEng degrees from Harbin Institute of Technology, China, and PhD degree from the Hong Kong University of Science and Technology, Hong Kong SAR. He is currently an Associate Professor at the College of Computing and Data Science, Nanyang Technological University, Singapore. He authored and co-authored more than 180 research papers in peer-reviewed journals, conferences, and books. His research interests include embedded and real-time systems, multiprocessor systems, network-on-chip, and machine learning acceleration.
\end{IEEEbiography}


\begin{IEEEbiography}[{\includegraphics[width=1in]{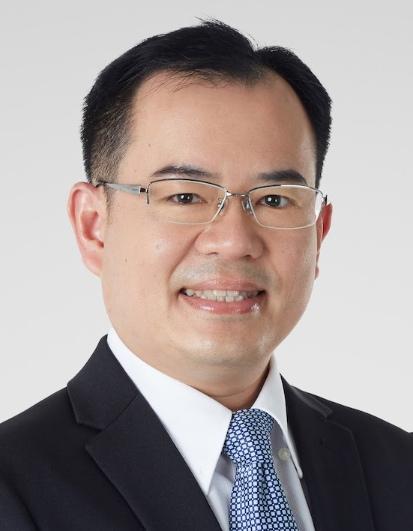}}]{Rick Siow Mong Goh}
received the Ph.D. degree in Electrical and Computer Engineering from the National University of Singapore, Singapore.
Associate Professor (Adj.) Rick Goh is Director of Computing \& Intelligence at A*STAR’s Institute of High Performance Computing (IHPC), Associate Professor (Adj.) at Duke-NUS Medical School, Co-Director of A*STAR-EVYD Joint Lab, Senior Principal Investigator (Adj.) at Singapore Eye Research Institute (SERI), and Co-Director of SERI-IHPC Joint Lab. He represented A*STAR to co-organise the inaugural AI Health Summit in 2022 with SingHealth and Ministry of Health. Rick has co-authored 150+ peer-reviewed papers in renowned clinical journals such as Nature Aging, Nature Genetics, The Lancet Digital Health, and top-tier AI and computing journals and conferences such as Nature Machine Intelligence, Nature Communications, IEEE TPAMI, TNNLS, TPDS, Computers, Cybernetics, Transactions on Medical Imaging, Medical Image Analysis, CVPR, CACM, AAAI, IJCAI, MICCAI, and Supercomputing Conference (SC). He has recently won multiple highly-competitive AI Singapore Tech Challenge and Grand Challenge grants, best paper awards, Healthcare AI project awards, and has been recognised with a National Award (COVID-19) Commendation Medal.
\end{IEEEbiography}


\begin{IEEEbiography}[{\includegraphics[width=1in]{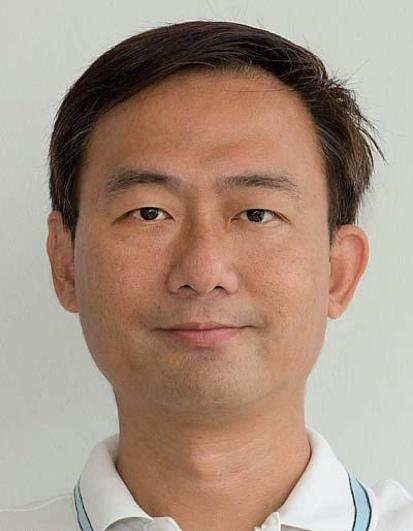}}]{Weng-Fai Wong}
received the BSc degree from the National University of Singapore, in 1988, and the DrEngSc degree from the University of Tsukuba, Japan, in 1993. He is currently an associate professor with the Department of Computer Science at the National University of Singapore. His research interests include computer architecture, compilers, and systems for machine learning. He is a senior member of the IEEE.
\end{IEEEbiography}

\end{document}